\documentclass[preprint,12pt]{elsarticle}

\usepackage[inline]{enumitem}

\usepackage{graphicx}
\graphicspath{{/graphs/}}
\DeclareGraphicsExtensions{.pdf,.jpeg,.png}
\usepackage{caption}
\usepackage{subcaption}

\usepackage{algorithm}
\usepackage{algpseudocode}

\usepackage{tabularray}
\UseTblrLibrary{booktabs}
\UseTblrLibrary{counter} 

\usepackage{url}
\usepackage{hyperref}
\hypersetup{colorlinks=true, linkcolor=blue, citecolor=blue,}

\usepackage[dvipsnames]{xcolor}
\usepackage{soul}

\usepackage{amsmath, amssymb, amsthm, gensymb}
\usepackage[utf8]{inputenc}
\usepackage{mathtools}

\usepackage{multicol}

\usepackage{acro}

\DeclareAcronym{AI}{
  short = AI,
  long = artificial intelligence}
\DeclareAcronym{DL}{
  short = DL,
  long = deep learning}
\DeclareAcronym{ML}{
  short = ML,
  long = machine learning}
\DeclareAcronym{RL}{
  short = RL,
  long = reinforcement learning}

\DeclareAcronym{NN}{
  short = NN,
  long = neural network}
  
\DeclareAcronym{ANN}{
  short = ANN,
  plural= s,
  long = artificial neural network}
  
\DeclareAcronym{RNN}{
  short = RNN,
  long =recurrent neural network}
  \DeclareAcronym{CNN}{
    short = CNN,
    long =convolutional neural network}
\DeclareAcronym{LSTM}{
  short = LSTM,
  long = long short-term memory
}

\DeclareAcronym{DNN}{
short=DNN,
long=deep neural network
}

\DeclareAcronym{AOI}{
short=AOI,
long=area of interest,
long-plural-form=areas of interest
}

\DeclareAcronym{ECMWF}{
short=ECMWF,
long=european center for medium-range weather forecasts
}

\DeclareAcronym{cv-mode}{
    short = GPH,
    long = Global Plant Holdout
  }
    
\DeclareAcronym{kn-mode}{
    short = GPkN,
    long = Global Plant kNear
    }
 
\DeclareAcronym{nRMSE}{
    long=normalised root mean square error,
    short=nRMSE,
    short-format =\textrm,
    extra= An error metric
}   
\DeclareAcronym{Serror}{
    long=forecast skill,
    short= S,
    short-format =\textrm,
    extra= An error metric
}

\DeclareAcronym{GBT}{
  short = GBT,
  long = gradient boosted tree}
\DeclareAcronym{XGB}{
  short = XGB,
  long = XGBoost}
\DeclareAcronym{LR}{
  short = LR,
  long = linear regression}

  \DeclareAcronym{KNN}{
    short = k-NN,
    long = $K$-nearest neighbours}

\journal{arxiv}

\begin{document}

\begin{frontmatter}



\title{Local-Global Methods for Generalised Solar Irradiance Forecasting}


\author[inst1]{Timothy R. Cargan}

\affiliation[inst1]{
organization={Computational Optimisation and Learning (COL) Lab, School of Computer Science},
            addressline={University of Nottingham}, 
            city={Nottingham},
            postcode={NG8 1BB}, 
            country={United Kingdom}}

\author[inst1]{Dario Landa{-}Silva}
\author[inst1,inst2,inst3]{Isaac Triguero}

\affiliation[inst2]{
organization={Department of Computer Science and Artificial Intelligence},
            addressline={University of Granada}, 
            city={Granada},
            country={Spain}
            }
\affiliation[inst3]{
organization={DaSCI Andalusian Institute in Data Science and Computational Intelligence},
            addressline={University of Granada}, 
            city={Granada},
            country={Spain}
            }

\begin{abstract}
As the use of solar power increases, having accurate and timely forecasts will be essential for smooth grid operators. 
There are many proposed methods for forecasting solar irradiance / solar power production.
However, many of these methods formulate the problem as a time-series, relying on near real-time access to observations at the location of interest to generate forecasts.
This requires both access to a real-time stream of data and enough historical observations for these methods to be deployed. 
In this paper, we propose the use of Global methods to train our models in a generalised way, enabling them to generate forecasts for unseen locations.
We apply this approach to both classical ML and state of the art methods.
Using data from 20 locations distributed throughout the UK and widely available weather data, we show that it is possible to build systems that do not require access to this data. 
We utilise and compare both satellite and ground observations (e.g. temperature, pressure) of weather data.
Leveraging weather observations and measurements from other locations we show it is possible to create models capable of accurately forecasting solar irradiance at new locations.
This could facilitate use planning and optimisation for both newly deployed solar farms and domestic installations from the moment they come online.
Additionally, we show that training a single global model for multiple locations can produce a more robust model with more consistent and accurate results across locations. 
\end{abstract}




\end{frontmatter}





\section{Introduction}

Power generators must accurately forecast their power output as any unplanned deviation put on the grid can push supply and demand away from equilibrium.
In order to maintain stability, the grid operator is forced to intervene, taking action to balance the grid.
The cost of this action, the balancing cost, is often passed onto the offending producer~\cite{UKGridDocs}.
As renewable energy sources become more prevalent, due to their variability, having timely and accurate forecasts of production is vital for their effective use~\cite{jones_renewable_2014}.
Predictions of just a few hours into the future can be used by generators to adjust their plans. 
In the case of the UK, generators can change their stated output (or by power) up to 1 hour before the generation period.

In the case of solar generation, irradiance (the power per unit area radiated from the sun~\cite{clearSky}) is converted into electricity. 
Power output typically tracks the sun, peaking in the middle of the day.
However, the amount of energy produced is dependent on total irradiance falling onto the panels and is susceptible to atmospheric interference~\cite{eeBook}.
Changes in weather conditions can cause output to fluctuate abruptly throughout the day.

In the literature, a wide array of techniques have been used for forecasting solar irradiance, from ARIMA and Support Vector Regression
\cite{Yang_Hourly_2012, Alrashidi2021} to \ac{DL} methods~\cite{solarIradRNN, Pan_Prediction_2021}.
However, many of these approaches depend on observations of irradiance at the \ac{AOI}.
For these methods to provide forecasts in ``production'', in addition to requiring enough historic data to effectively train a model, a feed of data from the \ac{AOI} is required.
Whilst new installations can be fitted with equipment capable of relaying irradiance observations, retrofitting existing plants can be cost-prohibitive, with domestic installations presenting an even larger challenge.

A solution to overcoming this data dependency is generating irradiance forecasts by regressing over weather data, a major factor in the variability of PV output.
Doing so can also uncouple predictions from real-time irradiance observations.
Access to point-based weather data (e.g. temperature, wind speed at a single location) is available from commercial providers who claim to offer accurate observations for virtually any point on the planet~\cite{weatherbit, openweather}.
This approach has been used successfully by~\cite{solarIradUsingForecast, solarUsingWeather, Omar_2022_input_selection}.
Additionally, real-time satellite imagery from providers such as~\cite{eumetsat} provide an alternative way to potentially capture a richer weather state as demonstrated in~\cite{Gallo_Solar_2022}.
The images capture features such as cloud cover; using a sequence of images, information such as their movements can be extracted and used in the forecasting models.

It has also been noted in the literature~\cite{solarMultiLoc, Lago2018} that most existing methods focus on providing forecasts for a single location.
Using this ``Local'' approach, each \ac{AOI} would require its own model with enough corresponding historic data to be trained.
When dealing with multiple \acp{AOI} developing and maintaining a Local model for each is not practical.
To say nothing of the challenges this approach would present if used on a domestic level.

For new installations, this data simply might not exist. 
Furthermore, applying this approach on a domestic level would result in potentially thousands of models, an outcome that seems fundamentally floored.

Rather than focusing on a single \ac{AOI} one can attempt to generate forecasts for multiple locations.
Taking this ``Global'' approach and creating a generalised model to predict for multiple \acp{AOI} eliminates the practical challenges of managing multiple Local models.
Furthermore, the use of a Global model presents several advantages. 
A Global model can result in higher quality forecasts by learning from multiple locations' data~\cite{Pablo2021, Semenoglou2021}.
One can even use a Global model to generate forecasts for an unseen \ac{AOI}.
This enables forecasting for locations, regardless of whether historical data is available, such as in the case of a new \ac{AOI}.
In this case, as time goes on, data collected at the \ac{AOI} could be used to refine the model further.

A limited number of works exist exploring the use of a Global model to generate irradiance forecasts for multiple \acp{AOI}~\cite{Lago2018, Bottieau_cross_2022}.
Both works create Global models capable of generating robust forecasts for multiple locations.
In \cite{Lago2018} they compared a Global DNN to alternative Local methods while \cite{Bottieau_cross_2022} specifically focused Global models for \acp{AOI} with no historic data.
Neither explores the use of various possible input features (e.g., weather data, real-time irradiance) in both Local and Global approaches.


The main goal of this work is to provide practical industry applicable approaches useful for short to medium-range (intra-day) planning.
We compare Local and Global models using several typical \ac{ML} methods used for irradiance forecasting. 
We further extend the Global approach to circumvent some of the potential real-world data dependency issues and analyse their effect on performance; \acf{cv-mode} to handle a lack of historic data and \acf{kn-mode} to cope with missing real-time data.
Additionally, we compare the use of satellite images to point base ground weather data.

\begin{itemize}
   \item We show that Global models can leverage data from multiple locations for improved forecasting performance and they can generalise to unseen locations removing the need for historical data to predict at unseen \acp{AOI}. 
   \item We analyse the impact use of real-time irradiance has on forecasts and explore methods to uncouple irradiance predictions from real-time observations by substituting observations from nearby plants. 
    \item We compare a number of standard \ac{ML} methods commonly used for forecasting Irradiance (Random Forests, DNN, LSTM and CNN).
    
    \item We show that using rich weather data from satellites can produce better forecasts. 
\end{itemize}

We validate our proposed methods using irradiance measurements and corresponding weather data from 20 \acp{AOI} distributed across the UK (see \autoref{fig:eumetsat_example_images}).
We compare the performance of our models across four training schemes, Local, Global, \ac{cv-mode} and \ac{kn-mode}, outlined in \autoref{ch3:subsection:data_models}.

The remainder of this paper is structured as follows; \autoref{section:background} provides an overview of the literature.
\autoref{ch3:section:methods} describes our proposed methods while
\autoref{experframework} outlines our experimental framework as well as gives details of the data used.
In \autoref{results} we present our results and analyse the performance of our proposed methods.
Finally, \autoref{section:conclusion} concludes the paper.

\section{Background}\label{section:background}

In this section, we provide background information on the problem domain and provide an overview of the various \ac{ML} based methods used for irradiance forecasting.
We start by formalising the irradiance forecasting problem in \autoref{ch3:method:problem_def}.
In \autoref{background:weather} we give an overview of the different kinds of weather data. 
\autoref{bacgkorud:ml_methods} gives an overview of the various \ac{ML} based irradiance forecasting methods, discussing their advantages and their main caveats.


\subsection{Problem Definition}\label{ch3:method:problem_def}
In its simplest form, our goal at time step $t_0$ is to predict future irradiance values $\theta_{t_0} = 
[{i}_{t+1}, \ldots, {i}_{t{+\textrm{fh}}}]$, where $\textrm{fh}$ is our forecast horizon.
If we assume there exists a function that can map an input $X$ to irradiance forecasts $f(X_{t_i}) = \theta_{t_i}$ we can define a \ac{ML} problem to approximate the function $f$ using a model $m$ such that $m(X_{t_i}) \xrightarrow[]{} \hat{\theta_{t_i}}$. 
Here $X$ represents any data that could be used by a model such as weather data, historic irradiance observations, or even the time of day.

\subsection{Weather Data}\label{background:weather}
There are numerous types of weather data. 
We focus on two types, ground-based and satellite weather.
Ground-based weather data is comprised of observations of a number of variables such as wind speed, temperature, etc, made at an observation station in a fixed location.
In the UK there exists a large fleet of stations distributed all over the country.
Both observations and forecasts for these locations are made easily accessible from commercial suppliers such as~\cite{weatherbit}.
Satellite data consists of observations of several wavelengths of light and is typically presented as an image where each pixel covers an area in the order of $3km^2$.
The satellite observations can be further processed to extract estimates of weather state such as cloud cover or temperature. 
Satellite images covering the UK are made hourly by~\cite{eumetsat}.

\subsection{\Acl{ML} methods}\label{bacgkorud:ml_methods}

In the literature, one can find multiple \ac{ML} based methods of irradiance forecasting.
These can be loosely divided into two methods:
regression-based (\autoref{background:regression}) and time-series based (\autoref{background:time_series}).
However many newer methods combine elements from both approaches (\autoref{background:hybrid}).

\subsubsection{Regression}\label{background:regression}
Regression models use correlated values to make predictions.
As previously noted, the predominant variable\footnote{Ignoring the effects of time of day, day of the year, and location} in how much solar irradiance makes it to the ground is the weather~\cite{jones_renewable_2014}.
Since weather and irradiance are correlated, a regressive model can be learned to predict irradiance using weather data.
Using the above notation, we formulate the problem as; given a set of weather values, $W_{t_i}=[wind, temp, pres]$;
we aim to create a model where $m(W_{t_i}) \xrightarrow{} \hat{\theta_{t_i}}$.

There are many examples throughout the literature of irradiance forecasts being created from regressing over weather data using a variety of techniques~\cite{solarReview, review_2020, weatherStationIrrad, Omar_2022_input_selection, Garcia_2022_combination, Alcantara_2023_deep_neural}.
Although classical \ac{ML} approaches such as Support Vector Regression and Decision Trees are actively used for irradiance forecasting~\cite{ solarIrradSatelliteSVM, Niu2020}.
In recent years, \acp{NN} have proved to be a highly effective tool for regression due to their power and flexibility as function estimators~\cite{Goodfellow-et-al-2016}.
Accordingly, a number of \ac{NN} based methods have been employed~\cite{solarUsingWeather, review_2020} with \acp{CNN} being used to regress over satellite images to produce predictions~\cite{Gallo_Solar_2022}.

An advantage of modelling irradiance predictions as a regression problem using weather data is that predictions are uncoupled from real-time irradiance observations such as in~\cite{Lago2018}.
However, regressive models typically map one set of input features to one output prediction~\cite{solarUsingWeather, solarIradUsingForecast}.
In order to forecast irradiance, future weather states are needed.
As such, regressive methods must rely on the use of externally provided weather forecasts.

While the use of weather forecasts can introduce another level of uncertainty to the models.
It also presents another challenge as updated predictions can only be made when new weather forecasts are provided.
As such, the forecast used can limit the timeliness of a model's predictions.
For example, If a model produces an hourly forecast with a 12-hour horizon using weather data that updates once every 6 hours, while predictions are made for all steps, there are times when the forecast is stale and its useful horizon reduced. For example, if a prediction was made at 6 am, covering all steps until 6 pm, by 9 am there are only 9, dated, forecast steps remaining.
As such, weather data sources used must be carefully considered as they can affect both the performance and capability of models.


\subsubsection{Time series}\label{background:time_series}
Time-series modelling is an alternative method for building forecasting models.
A time series is a sequence of data where the order of observations matters, typically there exists a sequential relationship between examples~\cite{RNNReview}.
Time series forecasting methods use previous observations of the value the model intends to predict as input.
In the case of our irradiance forecasting problem, observations from the last few hours would be used to predict the future values in the sequence.
Using the notation from above we can formalise the \ac{ML} problem as:
given a sequence of irradiance measures, $I_t = [i_{t_{-\infty}}, \ldots, i_{t}]$;
We aim to create a model $m$ such that $m(I_t) \xrightarrow[]{} \hat{\theta_{t_i}}$.

There are numerous methods are used throughout the literature for time series forecasting on a wide array of problems~\cite{liu_nonpooling_2019, passalis_deep_2019, tran_temporal_2019, Alassafi_Time_2022, Weerakody2021}.
We split these further into two approaches, autoregressive and sequence modelling.

\paragraph{Autoregressive} approaches use a lagged window over a number of the previously seen examples to create a forecast~\cite{MLBook}.
There are several examples of purely autoregressive models being used for irradiance forecasting.
These include classical statistical methods such ARIMA~\cite{Yang_Hourly_2012}, as well as \acp{DNN}~\cite{solarIradANN, reikard_predicting_2009}.
The length of the window is a hyperparameter that can be tuned, however, the computational complexity will increase accordingly.
While effective, autoregressive methods can only model sequences that fall within the lagged window. 
As such, they can fail to capture relationships that are beyond the length of the window~\cite{RNNReview}.

\paragraph{Sequence modelling} can solve this issue, by modelling an evolving state.
With the rise of deep learning, \acp{RNN} have become a popular way to do so.
In order to create forecasts, \acp{RNN} process elements sequentially.
At each step, $t$, taking both input features $X_t$, and the models' previous state $\alpha_{t-1}$ from the preceding step as input, and outputs a prediction, $\hat{\theta_{t}}$ and creating new state $\alpha_t$.

By passing in its previous state information can be transmitted from one time-step to another.
\ac{LSTM} architectures have proved to be a highly effective form of \ac{RNN}, by selectively taking in its state from previous time steps they are able to model sequence dependencies~\cite{RNNReview,lstmSpaceOdyssey}.
Some examples of \acp{LSTM} being used to generate irradiance forecasts can be found in~\cite{solarIradRNN, Pan_Prediction_2021}.

Despite their effectiveness, both autoregressive and sequence modelling approaches require access to real-time data in order to make predictions. From a practical standpoint, this limits the applicability of time series models to locations that can provide a feed of data.

\subsubsection{Hybrid}\label{background:hybrid}
While it is possible to use just the irradiance sequence or weather data to create a forecasting model. 
It is possible to create a hybrid architecture that can leverage both kinds of data in a single model.
Li et al proposed combing an ARIMA-based model with extra weather data to improve performance~\cite{Li_argmax_2014}. 
Many of the more recent methods we have classed as time series are in fact hybrid, using both irradiance and weather data~\cite{Pan_Prediction_2021, Alassafi_Time_2022}.

\subsection{Local and Global Models}\label{background:local_global}
Local and Global methods are ways to create forecasting models with datasets containing multiple time series.
The Local approach creates a model per series while the Global fits a single model to all of them~\cite{Pablo2021}.
In \cite{Semenoglou2021, Bottieau_cross_2022}, the Global approach is referred to as cross-learning.
In the context of irradiance forecasting, Local and Global approaches are different ways to manage the data and model(s) when there are multiple \acp{AOI}.
A Local model is an individual model learned for a specific location (\ac{AOI}).
A Global model is generalised across locations.
In our case, a Global model is defined as a single model that can predict all \acp{AOI} within the set.

While works such as~\cite{Zambrano2020} enable predictions for locations with limited history by selectively extracting extra data from correlated locations, they are still Local models and as such would struggle to generalise to unseen locations.
As previously mentioned, to the best of our knowledge a limited number of works exist on Global methods for irradiance forecasting.
In~\cite{Lago2018}, by leveraging a combination of satellite observations and irradiance forecasts from the \Ac{ECMWF} a Global DNN model was trained.
However, using the \ac{ECMWF} their ability to output timely forecasts is limited as their forecasts update every 6 hours.
To address these issues, we propose making use of standard, and widely commercially available, weather forecasts.
Many of these update sub-hourly removing any data dependency~\cite{weatherbit}.

Bottieau et al use weather forecasts to create a number of Global models using a number of \ac{ML} methods to generate predictions at locations with no historic data~\cite{Bottieau_cross_2022}.
Whilst their approach allows for predictions in locations without real-time irradiance observations, 
they achieve this simply by not including them.
We explore this as well as alternative methods such as \ac{kn-mode} that circumvent the data dependency by substituting values.

\section{Motivation and Methodology}\label{ch3:section:methods}

In this section we discuss our proposed methodology, outlining how we address the irradiance forecasting problem defined in \autoref{ch3:method:problem_def}.
In \autoref{ch3:method:subsection:motivation} we outline our motivations while,
\autoref{ch3:subsection:data_models} describes our proposed data models to solve the issues outlined.

\subsection{Motivation}\label{ch3:method:subsection:motivation}

Our goal is to present plausible methods, capable of being deployed in an industry setting that provided timely, accurate forecasts of irradiance.
To that end, our model must:
\begin{enumerate*}[label={(\arabic*)}]
    \item Produce forecasts at a frequency and with a forecast horizon to be of practical use;
    \item Generalise across locations;
    \item Uncouple irradiance forecasts from real-time observations.
\end{enumerate*}
As such, when designing our methods, we must take into consideration the various possible technical limitations when selecting input features, i.e. availability, update frequency, timeliness, etc.


Given these objectives and limitations, for our study, we aim to create models capable of predicting with an hourly frequency with a forecast horizon of 6 hours.
We show that predictions of steps after 6 hours are predominately dominated by the weather data and as such in ``production'' will be limited by weather forecast accuracy.


    



\subsection{Data Models / Training Schemes}\label{ch3:subsection:data_models}

The data available at both training and inference times dictate the overall design of any forecasting model.
Broadly, we consider two classes of data when designing the models:

\begin{enumerate}
    \item Historic observations - this is data that is guaranteed to be available at training time. This would be a database of weather and irradiance values for one or more \acp{AOI} spanning multiple years.
    
    \item Real-time - the data used to make the forecasts. It consists of telemetry/observations transmitted, in a timely manner (on the order of minutes), to the model in order for it to produce useful forecasts. 
    This could be an observation of irradiance measured at given \ac{AOI} or weather forecasts sourced from 3rd parties.
\end{enumerate}

Both classes of data are needed to create a useful forecasting model. However, for any given \ac{AOI} there may be limitations on data availability.
Using the problem definition in \autoref{ch3:method:problem_def}, we outline four ways to train a forecasting model using various combinations of possibly available data.
Each represents a conceivable dataset that could be available for a group of \acp{AOI} when attempting to build a forecasting model.

\begin{figure}[ht!]
\begin{subfigure}[]{0.5\textwidth}
\centering
    \includegraphics[width=0.95\textwidth]{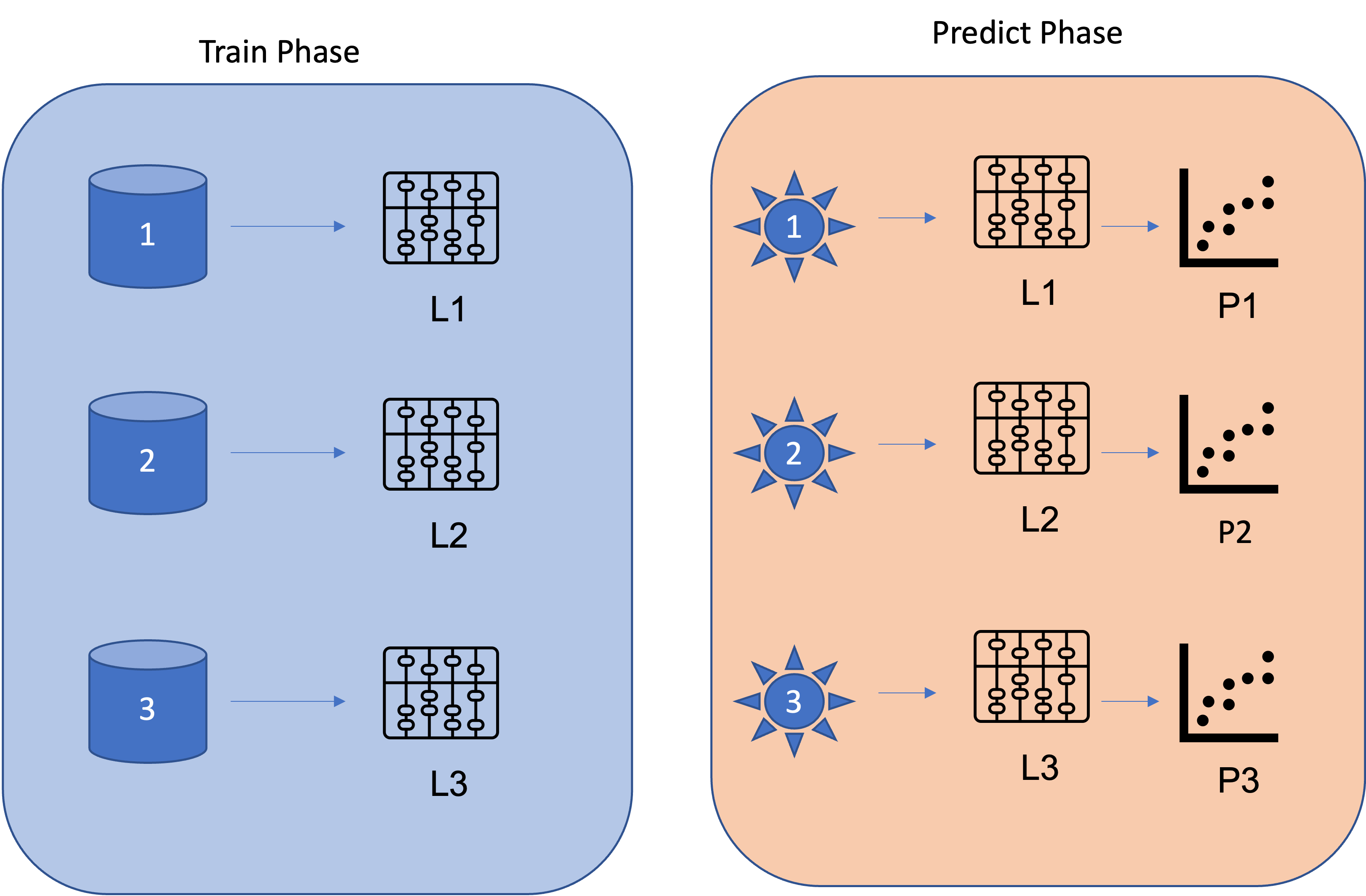}
    \caption{Local training}
    \label{fig:lgck_methods:local}
\end{subfigure}
\begin{subfigure}[]{0.5\textwidth}
\centering
    \includegraphics[width=0.95\textwidth]{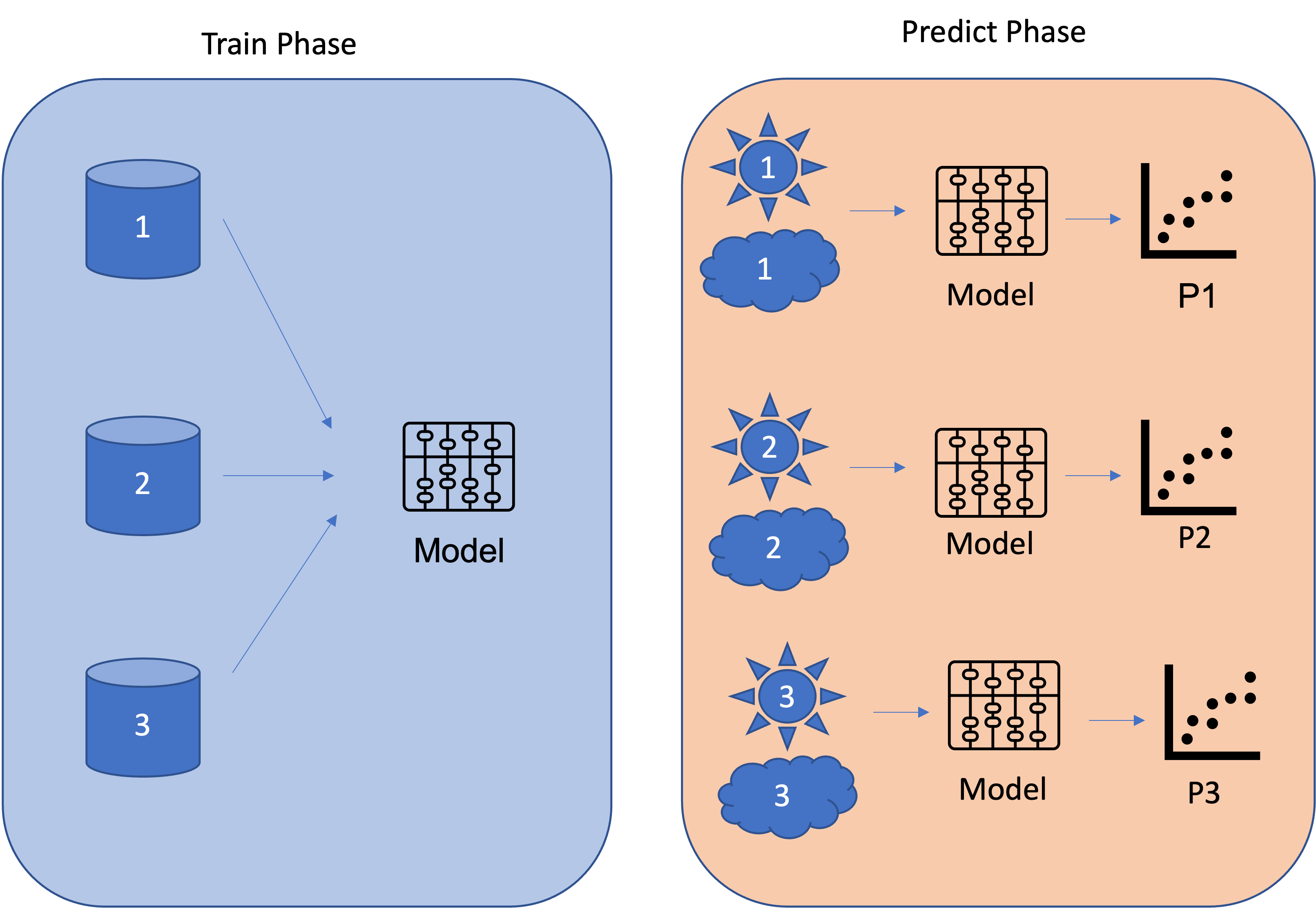}
    \caption{Global training}
    \label{fig:lgck_methods:global}
\end{subfigure}

\begin{subfigure}[]{0.5\textwidth}
\centering
    \includegraphics[width=0.95\textwidth]{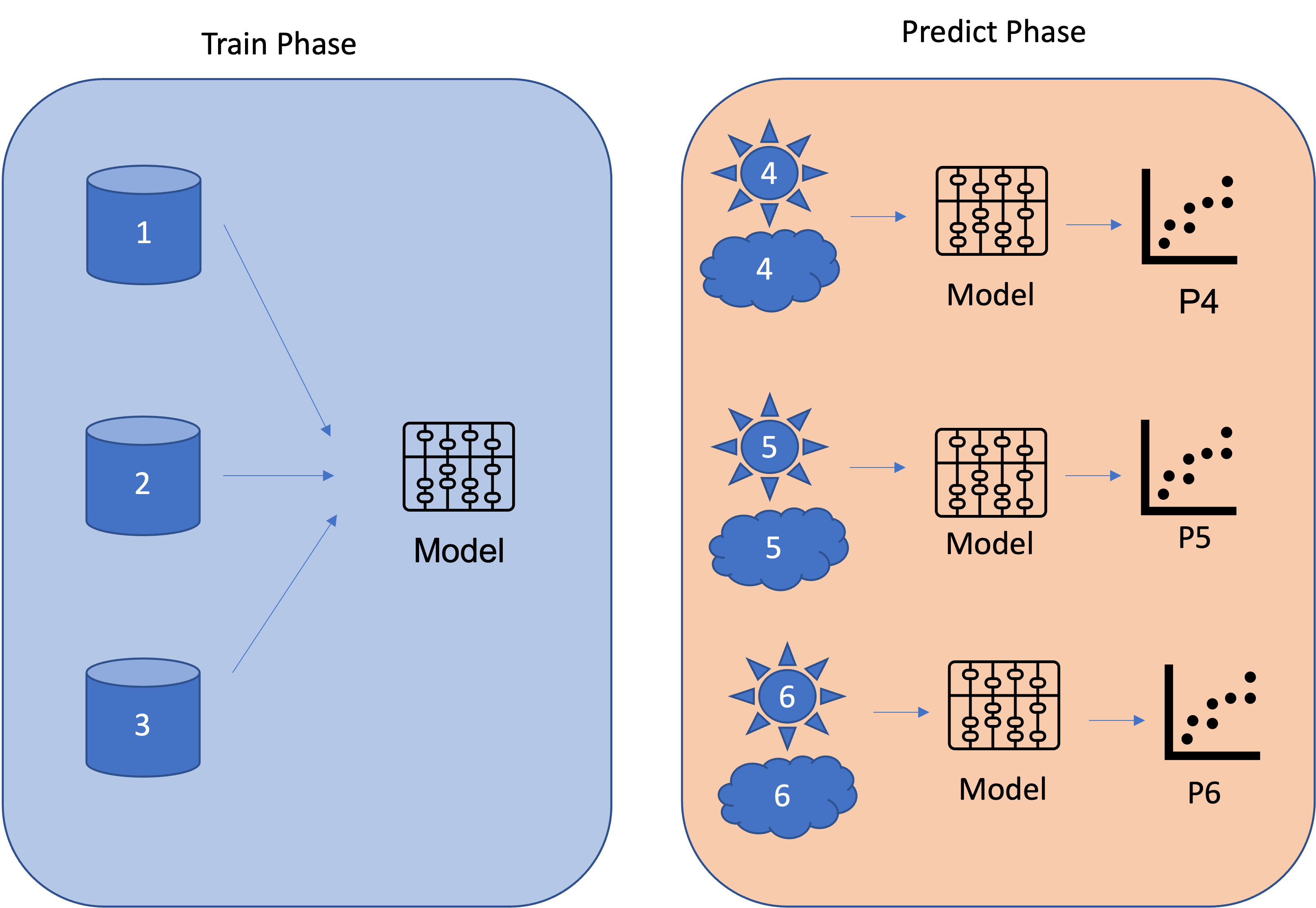}
    \caption{\ac{cv-mode} training}
    \label{fig:lgck_methods:cv}
\end{subfigure}
\begin{subfigure}[]{0.5\textwidth}
\centering
    \includegraphics[width=0.95\textwidth]{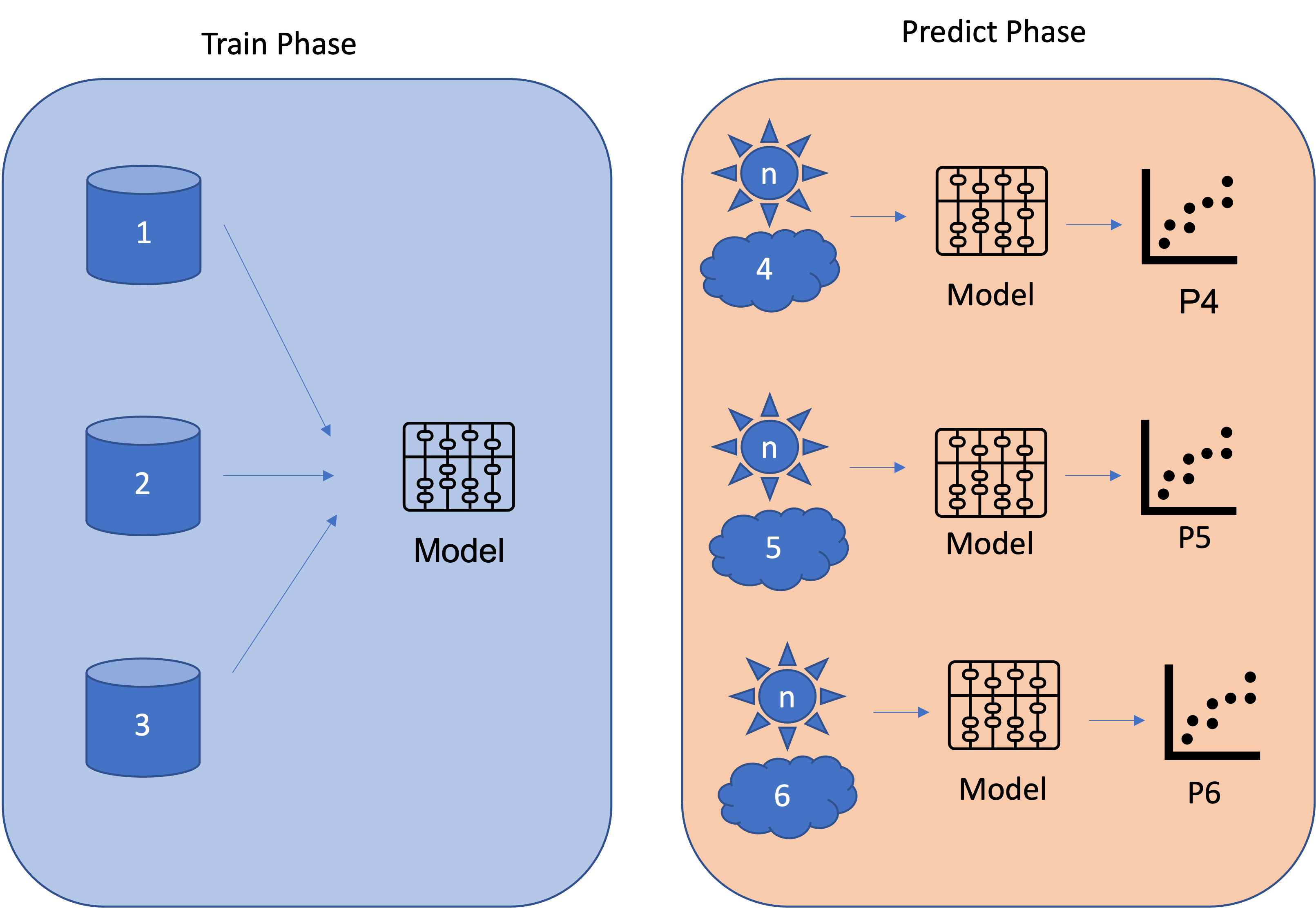}
    \caption{\ac{kn-mode} training}
    \label{fig:lgck_methods:kn}
\end{subfigure}

\caption{Training and predict data flows for the proposed training schemes}
\label{fig:lgck_methods}
\end{figure}

We use four data models, Local, Global, \acf{cv-mode} and \acf{kn-mode}.
Local is a typical approach with Global the logical way to generalise across locations solving some of the limitations of a Local model.
\ac{cv-mode} and \ac{kn-mode} are further extensions of the Global approach each solving a data limitation. An overview of the four methods is shown in \autoref{fig:lgck_methods}.

In this context, an \ac{AOI}s data consists only of irradiance values, it is assumed that weather data (ground-based or satellite) will always be available, both historically and in real time, for any \ac{AOI}s location.

\subsubsection{Local Models}
Local Models, as their name suggests, are localised to a single \ac{AOI}.
As such they are only trained on data from and produce predictions for a given \ac{AOI}.
This approach is common in the literature~\cite{review_2020} and would be a typical approach when there are a small number of \acp{AOI}.
Since local models are created by fitting a model using only data from the given \ac{AOI}, one model must be created per \ac{AOI}. 
While being a relatively straightforward approach Local models have three key limitations:
\begin{itemize}
    \item They require enough historic data for each \ac{AOI} to effectively learn a model (the amount of historic data needed is dependent on the model and data being used).
    \item A model per \ac{AOI} must be generated. If there are a large number of \acp{AOI}, such as in the case of a domestic solar fleet, computation and storage constraints could become an issue.
    \item If used, a real-time data feed of irradiance for all \acp{AOI} may be needed.
\end{itemize}

\subsubsection{Global Model}
In the case of the Global model, we create a generalised version of the local method, a single model that can predict for all \acp{AOI}~\cite{Lago2018}.
To do so, we train a single model on data for all \ac{AOI}s.
This is done by unioning all data for every \ac{AOI} into a single composite dataset and using it to train a model.
Global models solve two of the issues presented by local models. 
\begin{enumerate*}[label={(\arabic*)}]
    \item A single model artefact is created, greatly reducing the overhead of managing multiple models,
\item Since they are trained on multiple \ac{AOI} it is possible to create an effective model even if some \ac{AOI}s have limited historical data.
\end{enumerate*}

This definition of a Global model still assumes the perfect case where there is access to at least partial historic data for all \ac{AOI}s, and if used, a real-time feed of irradiance for all \ac{AOI}s. 
Because of this, the Global approach is still, by definition, limited to \ac{AOI}s with historic data. 
Additionally, if an \ac{AOI} has missing real-time data no predictions can be made.

\subsubsection{\acf{cv-mode}}

We can extend the Global approach in an attempt to resolve these limitations. 
In the case of \ac{cv-mode}, we eliminate the need for all \ac{AOI}s to have historic data.
Like the standard Global approach, we aim to produce a single model able to effectively predict for several \ac{AOI}s, however not all \ac{AOI}s have historic data available.
This would be the case e.g. for a new installation or a new domestic \ac{AOI} without historic data.
This can be useful as data from decommissioned plants can still be used to train the models.

For \ac{cv-mode}, a single model is trained using the data from \acp{AOI} with available historic data (even if only partial). The model is then used to make predictions for all \acp{AOI} both with and without historic data, using their real-time data if needed. For this approach to be viable, any models created must be able to generalise well to new unseen \acp{AOI}.

To simulate this, a standard cross-validation approach is taken. Each \ac{AOI} is randomly assigned to one of 5 folds. At train time, for each fold a model is created using data from all but the \acp{AOI} in the given fold. At test time only the \acp{AOI} in the fold are evaluated.  This process is repeated for each fold resulting in a worst-case prediction for every \ac{AOI} in the training set.

\subsubsection{\acf{kn-mode}}
This approach only applies when the use of real-time irradiance is needed, \ac{kn-mode} attempts to solve the \emph{worst case} scenario where we have neither access to historic irradiance measures nor real-time data for the \ac{AOI}. 
While an unlikely scenario, it sets a baseline as the most challenging scenario. It could be a viable fallback strategy in the case of a sensor failure, used in a domestic setting, or to estimate the output an \ac{AOI}.  
For \ac{kn-mode}, we produce forecasts for multiple \ac{AOI}s where historic and real-time data is only available for a subset of the \ac{AOI}s. Like \ac{cv-mode}, a single model is trained using data from the \acp{AOI} with historic data. To generate predictions for the \ac{AOI}s with no data, we substitute the real-time values from the nearest \ac{AOI} with data.

To evaluate \ac{kn-mode}, the same per-fold \ac{cv-mode} model was used, however, the real-time irradiance values, used as an input feature, were substituted to the nearest plant not in the hold-out fold. 
Distances were calculated using the haversine function.


\section{Experimental Framework}\label{experframework}
In this section, we present our experimental framework. In
\autoref{ch3:subsection:data_details} we provide the details of the raw data used as well as any prepossessing that was done.
In \autoref{experfamework:model_conf} we describe the models used and their configurations.
Finally, \autoref{subsection:metrics_validation} explains the error metrics and validation methods used.

\subsection{Data Details}\label{ch3:subsection:data_details}
The raw data is comprised of data from three unique data sources all covering the period 2015-01-01T00:00 to 2021-01-01T00:00.

\paragraph{Irradiance data} We sourced irradiance values from the “MIDAS Open: UK hourly solar radiation data~\cite{irrad_dataset}. It consists of hourly irradiance ($ \frac{\textrm{KJ}}{\textrm{m}^2} $) from over 80 locations distributed throughout the UK. Each location consists of a time series with data for all or part of the period. We selected a subset of 20 locations to use as our primary \ac{AOI}s, the locations were selected as they have the fewest missing values for the timespan.
This was done to as fair as possible comparisons when evaluating local models between \acp{AOI}.

Pre-processing -- Any negative values in the dataset were replaced with a 0. The values were then transformed using the function $I = \max(3, \ln(\textrm{irradiance}+1)) - 4$. This was done to give an approximately normal distribution in the range $[-4, +4]$ centred on $0$ in the daytime hours, with raw values of less than $\approx20$ clipped out as we consider them night-time.

\paragraph{Weather data} Historic, hourly, ground station weather observations for all locations within the irradiance dataset. The data was sourced from a commercial supplier~\cite{openweather} and contains the features outlined in \autoref{table:point_weather_data_details}. The weather features were normalised using the methods outlined in the table.

\begin{table}[htbp]
\centering
 \begin{tblr}{c|c}
    \toprule
  \textbf{Feature} & \textbf{Normalisation} \\
  \midrule
     Cloud cover (\%) & Z-score Mean: 60, std: 30		\\
     Clear-sky irradiance ($\frac{W}{m^2}$) & $\ln$	\\
     Precipitation ($\frac{\textrm{mm}}{\textrm{hr}}$)  & Z-score Mean: 0.1, std: 0.33	\\
     Pressure (mb)	              & Z-score	Mean: 1000, std: 15.5 \\
     Relative humidity (\%)	      & Z-score	Mean: 82, std: 13	\\
     Temperature (\degree k)     & Z-score Mean: 283, std: 5.5	\\
     N/S component of wind	 &	- \\
     E/W component of wind	 & -	\\
\end{tblr}
\caption{Ground-based weather features used by the models.}
\label{table:point_weather_data_details}
\end{table}

\begin{figure}[ht]
\centering
    \includegraphics[width=7cm]{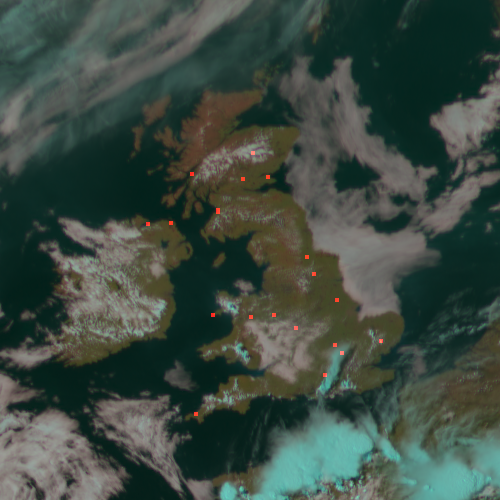}
    \caption{UK area of visible wavelengths processed to RGB with the locations of the \acp{AOI} highlighted in red}
    \label{fig:eumetsat_rgb_images}
\end{figure}

\begin{figure}[ht]
\begin{subfigure}[]{0.5\textwidth}
    \includegraphics[width=7cm]{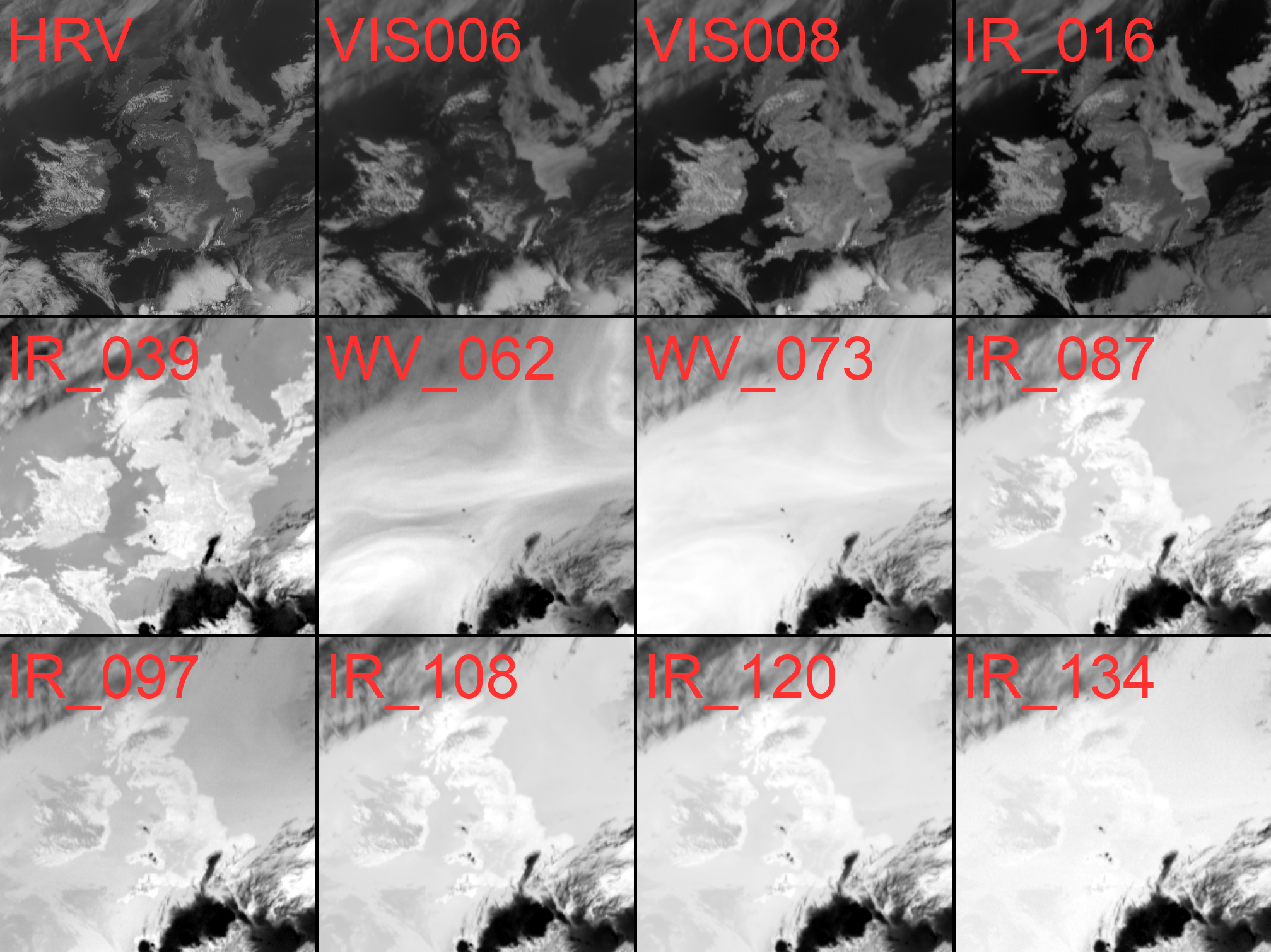}
    \caption{Day time satellite images (2018-05-28T15:00)}
\end{subfigure}
\begin{subfigure}[]{0.5\textwidth}
    \includegraphics[width=7cm]{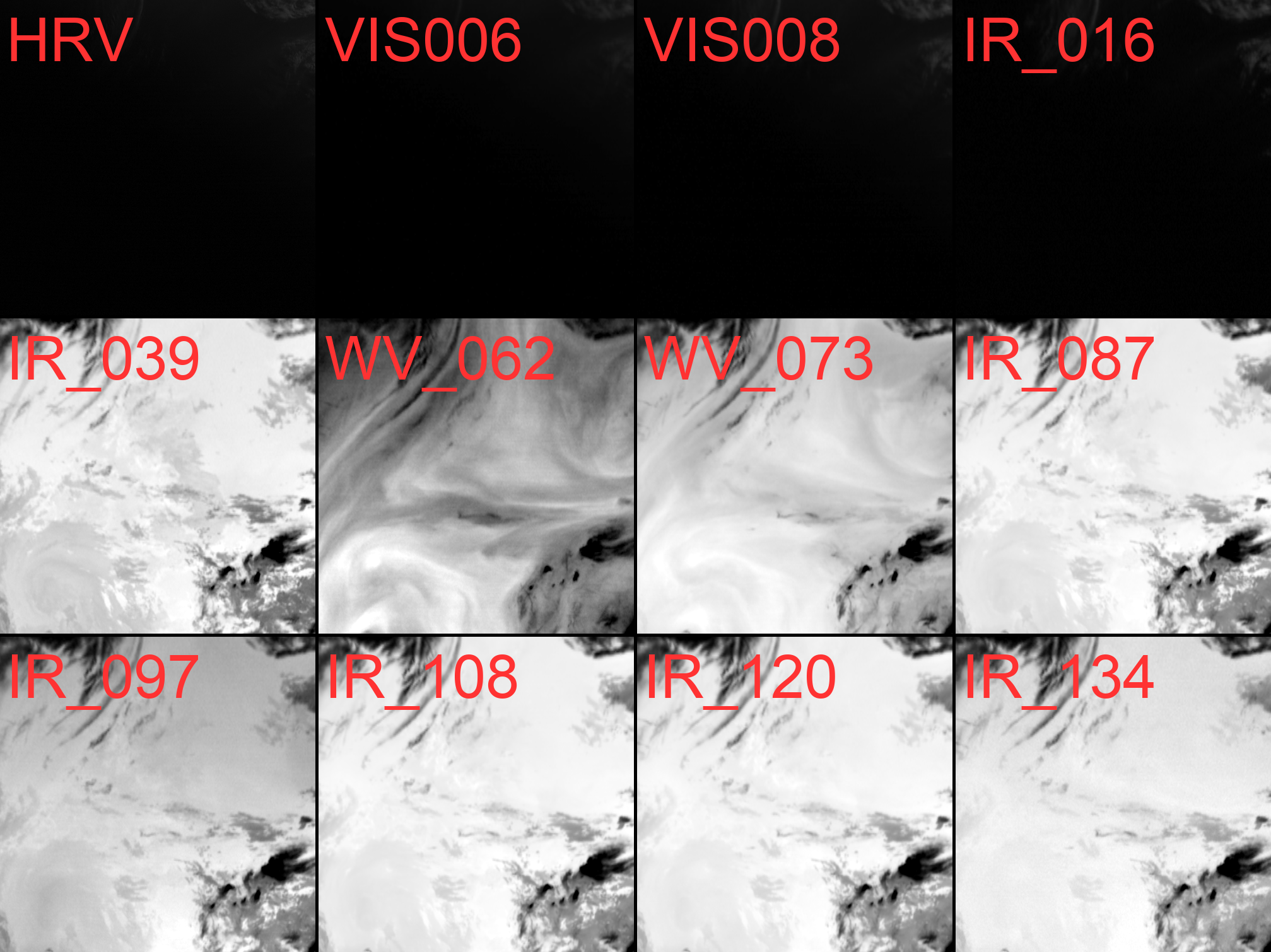}
    \caption{Night-time satellite images (2018-05-28T03:00)}
\end{subfigure}
\caption{A sample of the satellite images covering the full UK area}
\label{fig:eumetsat_example_images}
\end{figure}

\paragraph{Satellite Data} Hourly Satellite imagery from EUMETSAT~\cite{eumetsat}. 
The raw data is comprised of 11 channels spanning wavelengths from visible to infrared and an additional 12th channel in the visible spectrum at a higher resolution. We process a set of 12 images, 500px X 500px, of the UK area [-12.0W, 48.0S, 5.0E, 61.0N], with each pixel being approximately 3km by 3km.
The area processed is shown in \autoref{fig:eumetsat_rgb_images} with the location of \acp{AOI} highlighted. An example for each wavelength at day and night is shown in \autoref{fig:eumetsat_example_images}.

Pre-possessing –- The images are cropped so that they are centred on the given \ac{AOI}.

\paragraph{Calculated Values} In addition to the three data sources above, a fourth pseudo data source of values calculated from the \ac{AOI}s latitude, longitude and date time are used. The values calculated are outlined in \autoref{tab:caclulated_feats}.
These values help give the model context for when and where the forecast is being generated. 
As these values can be calculated simply from the \ac{AOI} location and time it is assumed they are always available.

\begin{table}[htbp]
    \centering
    \begin{tblr}{c|c}
    \toprule
    \textbf{Feature} & \textbf{Description} \\ 
    \midrule
Year sin  &  	$\sin(ts  * 2\pi/\textrm{Year Seconds})$\\
Year cos  & 	$\cos(ts  * 2\pi/\textrm{Year Seconds})$	 \\
Day sin   & 	$\sin(ts  * 2\pi/\textrm{Day Seconds})$	 \\
Day cos   & 	$\cos(ts  * 2\pi/\textrm{Day Seconds})$	 \\
Solar altitude & $\sin(\rm{Solar Elevation Angle})$\footnotemark{}\\
Azimuth sin & 		E/W of the position of the sun in the sky \\
Azimuth cos & 		E/W of the position of the sun in the sky \\
    \end{tblr}
    \caption{Calculated features, all solar positions were calculated using pvlib~\cite{pvlib}}
    \label{tab:caclulated_feats}
\end{table}

\subsection{Model Configurations}\label{experfamework:model_conf}

As mentioned in \autoref{ch3:method:problem_def} our aim is to create a model that approximates the function that maps its inputs to future irradiance values. 
There are many \ac{ML} methods that can be used to do this.
In order to evaluate the different data models outlined in \autoref{ch3:subsection:data_models}, and the various possible input features described in \autoref{ch3:subsection:data_details}; we utilise four typical \ac{ML} methods, Trees, \acp{LSTM}, \acp{DNN} and \acp{CNN}. 
While both the Trees and \ac{DNN} are purely regressive methods, the \ac{CNN} and \ac{LSTM} contain elements of both regressive and time series approaches.
Specifically, they contain explicit architectures to capture the sequential nature of the data and generate their forecasts autoregressively, using previous outputs as inputs to generate the next step.
All methods are able to make use of both the real-time irradiance and the calculated values as input features. 
However, the \acp{LSTM}, \acp{DNN}, and Trees use the ground-based weather data, while the \ac{CNN} uses the satellite data.

An effort was made to ensure hyperparameters and architectural decisions will produce results indicative of the approach’s possible performance, however, no full hyperparameter search was performed and better values or architectures may exist.
The same model configuration was used regardless of the data model or input features where possible.
The architecture and hyperparameters selected for each model are outlined below.

\subsubsection{Trees}
Trees, or tree-based models, are a classical ML approach. We use a random forest (RF), an ensemble model, training multiple decision trees each on subsets of the training data and combining the predictions from each tree to produce the final output~\cite{Breiman2001}.
For the rest of this article, we use the term Trees interchangeably with Random Forests.
Trees were chosen as they have been used effectively used on numerous forecasting problems. Their use of an ensemble approach makes them highly robust to over-fitting. They are also easy to implement with many standard library implementations.

The following hyper-parameters were used for all experiments: 
\begin{table}[h]
    \centering
    \begin{tblr}{ll}
        \toprule
        \textbf{Hyper Parameter}      & \textbf{Value} \\ 
        \midrule
        Number of Trees         & 20 \\
        Min Examples per Leaf   & 2 \\
        Max Depth               & 32
    \end{tblr}
    \caption{Hyperparameters used by the random forests}
    \label{tab:tree_hyperprams}
\end{table}

\paragraph{Forecasting}\label{subsubsection:indForecasting}
While it is possible for a tree-based model to output multiple labels\cite{Vens2008, Basgalupp2021}
A limitation of the implementation we used for our tree-based is that they can only output a single value.
This means that unlike the other \ac{DL} approaches used,
in order to produce forecasts a unique model was trained for each step in the forecast horizon $t_1 \ldots t_{fh}$.

\subsubsection{\Acf{DNN}}

The \ac{DNN} is a standard \ac{DL} method. 
A \ac{DNN} is typically comprised of an input layer, followed by a number fully connected hidden, and finally an output layer that produces the final result. 
Between each layer is a non-linear activation function such as Sigmoid or the ReLU. 
Each layer is comprised of several units, each of which outputs a weighted sum of all the previous layers' outputs. 
During the training phase, the weights are adjusted using backpropagation in order to minimise a loss function, such as mean square error in a regression context, with respect to training data.

In our case, the \ac{DNN} takes as input the ground-based weather data, past irradiance and the calculated features.
All inputs for every time step are stacked together as a single input vector.
The \ac{DNN} consists of 3 hidden layers all 128 units wide with a ReLU between each. The final output layer is forecast horizon units wide producing all 6 outputs at the same time.


\subsubsection{\Acf{LSTM}}

The \acp{LSTM} use the same input features as that of the \ac{DNN} and Tree models.
For the first 12 input steps, i.e. past observations where there are both weather and irradiance data, the \ac{LSTM} is in a warm-up phase establishing its internal state.
Once the past data has been consumed, the model enters the prediction phase, where for each prediction step $t_i$ the previous output of the model at $t_{i-1}$ is used in place of observed irradiance.

The topology of our \ac{LSTM} network is as follows:
The weather features are combined into a single vector and passed through a 3-layer MLP with 32 units. The output is then concatenated with the calculated features and passed through an \ac{LSTM} layer with 128 units. The output of the \ac{LSTM} layer is passed through another MLP with 2 hidden layers of 128 units with a final layer of 1 unit to produce the prediction.
Of note is that all MLP layers share their weights across every time step.


\subsubsection{\Acf{CNN}}

Unlike the other \ac{DL} based models, the \ac{CNN} use satellite images rather than ground-based weather data.
However, they can still incorporate both the real-time irradiance and calculated values as input.
The images for each time step are stacked into a 4D tensor of shape [timesteps, height, width, channels], resulting in a standard weather state input with a shape: (18,16,16,12).

The 4D tensor is then passed into a \ac{CNN} with a stack of three convolutional blocks made up of; a convolution with a kernel size of (3x3x3) convolving over both space and time with 16 features; A max pool of (1,2,2), reducing over just spatial dimensions; finally, a ReLU activation.

The output of this is treated as a weather feature vector like those used in the other methods, combined with the calculated features and passed into a model with the same architecture as that of the \ac{LSTM}.
Preliminary results suggested that the \ac{LSTM} performed better than the \ac{DNN} so was selected. 


\subsubsection{Training}
In practice, we only use the last $n$ values as irradiance values at step $t_0$ do not depend on historical values ad infinitum. 
Depending on the forecast problem a sensible value of $n$ must be selected, in our case, we have selected $n = 12$.

All the \ac{DL} models were trained with the setting specified in~\autoref{table:modelParams}.

\begin{table}[h]
  \centering
  \begin{tblr}{ll}
    \toprule
    \textbf{Hyperparameter}  & \textbf{Value} \\
    \midrule
    Learning rate       & $0.0003$ \\
    Loss                & $\textrm{MSE}$ \\
    Optimiser           & Adam \\
    Epochs               & $20$ \\

  \end{tblr}
\caption{No architectural hyperparameters common to all \ac{DL} models} \label{table:modelParams}
\end{table}

\subsection{Metrics and Validation}\label{subsection:metrics_validation}

There exist numerous metrics for evaluating the performance of regression models throughout the literature.
They primarily provide a summary of the error distribution where the error is defined as the difference between the observed and predicted value~\cite{cvForAutoRegresive, cvForTimeSerise}.
Additionally, there exist many ways to measure forecast accuracy~\cite{TimeSeriseError}. A common feature of forecast errors is a scaling of the error enabling comparison across distributions.

We define our error metrics as follows;
For a dataset comprised of $n$ examples where $Y = [y_0, \ldots, y_n]$ represents the observed value and $\hat{Y} = [\hat{y_0}, \ldots, \hat{y_n}]$ the predicted values.
The  following performance metrics defined below are used to evaluate and compare the various models:

\begin{itemize}

    \item \Acf{nRMSE}

    \begin{align}\label{eq:nRMSE}
         \textrm{nRMSE} &= \frac{\sqrt{\frac{1}{n}\sum_{i=0}^{n}{(y_i - {\hat{y}}_i)}^2}}{\overline{Y}}
    \end{align}
    \item \Acf{Serror}
    \begin{align}\label{eq:forecast_skill}
        \textrm{S}  &= \frac{1}{p}\sum_{w=0}^{p} \left( 1 - \frac{U_w}{V_w} \right)
    \end{align}
    Where: 
    The dataset is split into $p$ periods.
    $U_w$ represents the error of the models' forecasts for a given period, calculated as the mean square of the error scaled by the clear sky: $\sqrt{\frac{1}{n}\sum_{i=0}^{n}{(\frac{y_i-\hat{y_i}}{\textrm{GHI}}})^2}$.
    $V_w$ is a measure of the forecast difficulty for a given period calculated as the average of the variability of the irradiance $\sqrt{\frac{1}{n}\sum_{i=1}^{n}{(\frac{y_{i-1}}{{\rm GHI}_{i-1}} - \frac{y_{i}}{{\rm GHI}_{i}}) ^2}}$.
    
    $S$ is in turn calculated as the average of all the periods within the dataset. 
    We used a period size of 1 calendar month.

\end{itemize}

Both \ac{nRMSE} and \ac{Serror} are scale-invariant enabling a nuanced comparison between models and various output distributions. 
\ac{nRMSE} was selected as it has been widely used in the literature. While being scale-invariant, absolute errors are punished the same regardless of the size of the target, i.e. a prediction of $15$ for a true value of $10$ results in the same error as a prediction of $105$ for a true value of $100$. 
\ac{Serror} was proposed by~\cite{ForecastSkill} and is similar to mean absolute scaled error, adjusting in proportion to the size of the target sequence however the metric also factors in a measure of forecast difficulty.

\paragraph{A note on the error metrics and their interpretation}
\ac{Serror} and \ac{nRMSE} are interpreted in inverse of one another. 
In the case of \ac{nRMSE} – a lower value is better with $0$ indicating the predictions are perfectly accurate. A value of $1$ would mean forecasts are very bad as the $RMSE$ is equal to the mean of the sequence, as such the average absolute error is the same as the mean of the sequence. 
\ac{Serror}, conversely, is interpreted with a higher value indicating better performance. \ac{Serror} can fall in the range $(-\infty, 1]$ although a value less than $0$ indicates poor performance. A more detailed interpretation is in \autoref{tab:serror_interp}.

\begin{table}[h]
    \centering
    \begin{tblr}[]{c X[j,valign=m]}
    \toprule
     \textbf{Value} & \textbf{Interpretation} \\ 
    \midrule
    $1$ & The prediction is perfectly accurate. \\
    
    $0$ & The prediction is no better than that of a persistence model using the ratio of the last observed irradiance to clear sky
    $\hat{y}~=~\frac{I_{-1}}{{\rm GHI}_{-1}}{\rm GHI}_0$. \\
    
    $<0$ & Negative values indicate that the prediction is worse than the persistence model. \\
    \bottomrule
    \end{tblr}
    
    \caption{A description of \acf{Serror} interpretation}
    \label{tab:serror_interp}
\end{table}

nRMSE punishes errors equally regardless of the size of the target. As such, a good \ac{nRMSE} indicates that the magnitude of the errors is consistently small i.e. $\hat{y} = y \pm 30$. 

A low \ac{nRMSE} error but poor forecast skill could indicate that the model performs poorly when the target irradiance values are low, early and late in the day e.g. the model always predicts 50 above the true value. It could be due to a low variance in targets vs GHI making it `easy' to predict the sequence.
Conversely, a higher \ac{nRMSE} but good forecast skill could indicate that the models performed well during the day when target values are higher e.g. the model was always 10\% over the target value, or that the sequence was challenging to predict, leading to larger errors.

All errors presented are calculated using only daytime values. We defined daytime as any point where the target irradiance, $y > 20$ and $\rm GHI > 1$. We use both conditions to minimise the risk of any sensor errors sewing the results.

\subsubsection{Validation}

The data was split into train and test partitions of roughly 70\% train and 30\% validation. The data was split on 1st May 2019 with all models being trained on data from before the split date and evaluated on data after.
The date was arbitrarily chosen from a previously used dataset.

For both \ac{cv-mode} and \ac{kn-mode} we train each model on a subset of 16 \ac{AOI}s for historic data and only test on the missing 4. We use a standard cross-validation approach to generate predictions for all 20 \ac{AOI}s in our test set.

\subsection{Statistical Tests}
We use two non-parametric statistical tests for hypothesis testing to give statistical support when analysing our results~\cite{sheskin2003handbook}.
We use non-parametric tests as the initial conditions required for parametric tests to be reliable may not be met.
For pairwise comparisons, we make use of the Wilcoxon test~\cite{demvsar2006statistical, garcia2008extension}.
We assume a level of significance of $\alpha = 0.1$
To evaluate our methods against one another we use the Friedman Aligned-Ranks test~\cite{hodges1962rank} to identify statistical differences among them. We use the Holm post-hoc test to determine which algorithms have significant differences among the $1*n$ comparisons ~\cite{holm1979simple}.

\section*{A note on Weather Forecasts}
In order to evaluate our models we use historical observation in place of actual weather forecasts.
This was done to remove a degree of uncertainty caused by any error in the weather forecast as we attempt to understand the effect different input features can have.
As such the results presented are best case for any given method and in production using real weather forecasts we would expect a drop in performance depending on the accuracy of the forecasts.


\section{Analysis of Results}\label{results}

Here we present our results and analysis.
In \autoref{ch3:subsection:effect_of_input} we address what input features are the best.
Using a single forecasting model, we analyse the effects of various combinations of inputs on both the Local and Global approaches.
In \autoref{ch3:results:results_local_vs_global}, we explore the use of Local and Global methods, analysing the various performance of models trained with each scheme as well as the effects use of different kinds of weather data. 
Then, in \autoref{ch3:results:cv_kn} we show the performance of \ac{cv-mode} and \ac{kn-mode} training schemes to circumvent used to circumvent data limitations.
We also revisit the effects different input features can have specifically with a focus on uncoupling real-time irradiance.

\subsection{Effect of Inputs}\label{ch3:subsection:effect_of_input}
In this section we analyse the effect using different input features can have on model performance.
To do so we trained the \ac{DNN} on various combinations of inputs.
We focused on the \ac{DNN} due to their easy ability to change input features and their speed to train.
As outlined in \autoref{ch3:subsection:data_details}, there are a number of potential input features that could be used by the models. We group the various features into three classes
\begin{enumerate*}[label={(\arabic*)}]
    \item Calculated values such as location, time, solar position etc.
    \item Observations of the last few irradiance values.
    \item Weather features, past observations as well as forecasts of future states.
\end{enumerate*}
We trained the \ac{DNN} using four combinations of these inputs:

\begin{itemize}
    \item All - Irradiance, weather and static values
    \item Irradiance - Irradiance and static values
    \item Static - Calculated values
    \item Weather - Weather and static values

\end{itemize}

These four input combinations were selected to cover the range of possible data availability; from a full data set with both historic and real-time values for every \ac{AOI} to no data for any given \ac{AOI}. 

\begin{figure}[ht!]
\begin{subfigure}[b]{0.5\textwidth}
\centering
    \includegraphics[width=\textwidth]{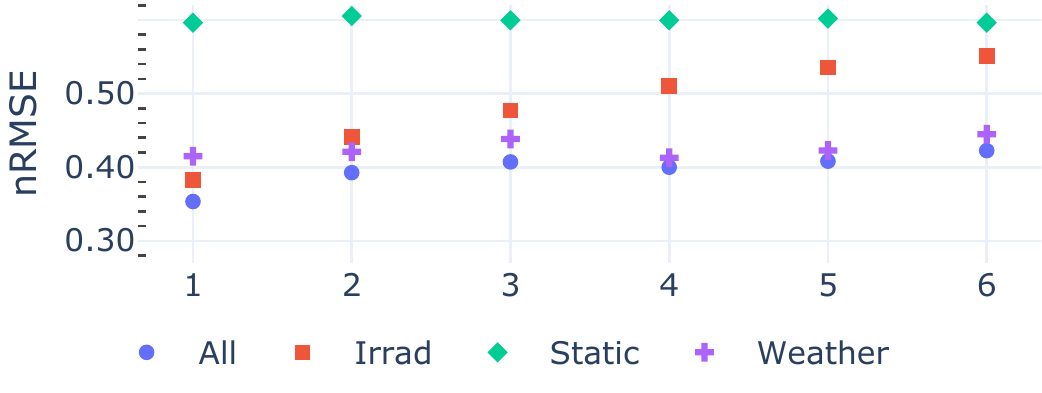}
    \caption{Local \ac{nRMSE} per step - lower is better}
    \label{fig:scatter_4mode_inputs_nRMSE_per_step_local}
\end{subfigure}
\begin{subfigure}[b]{0.5\textwidth}
\centering
    \includegraphics[width=\textwidth]{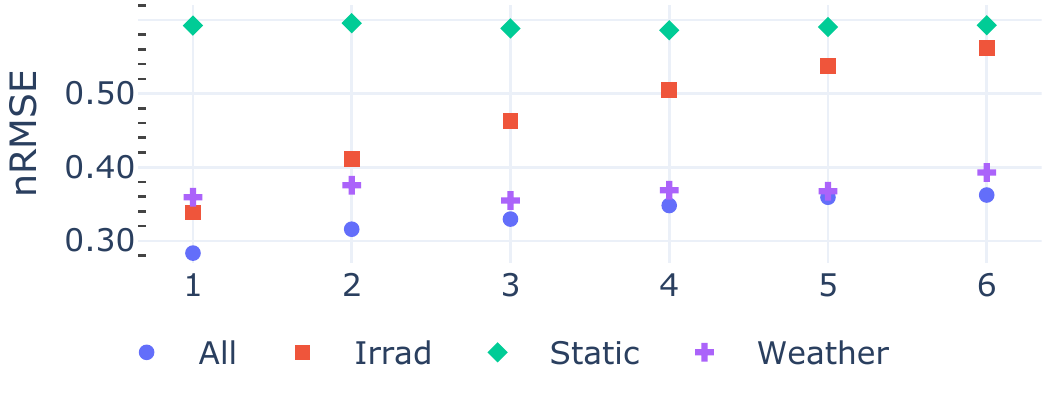}
    \caption{Global \ac{nRMSE} per step - lower is better}
    \label{fig:scatter_4mode_inputs_nRMSE_per_step_global}
\end{subfigure}
\begin{subfigure}[b]{0.5\textwidth}
\centering
    \includegraphics[width=\textwidth]{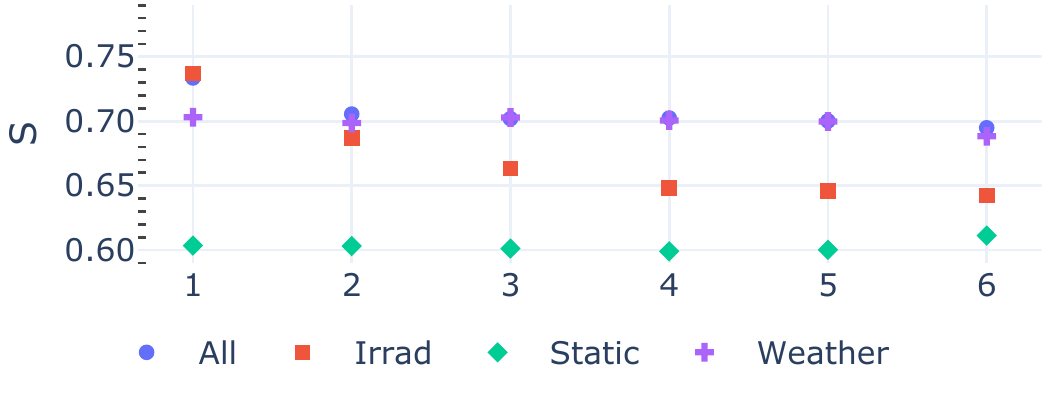}
    \caption{\ac{Serror} per step - higher is better}
    \label{fig:scatter_4mode_inputs_skill_per_step_local}
\end{subfigure}
\begin{subfigure}[b]{0.5\textwidth}
\centering
    \includegraphics[width=\textwidth]{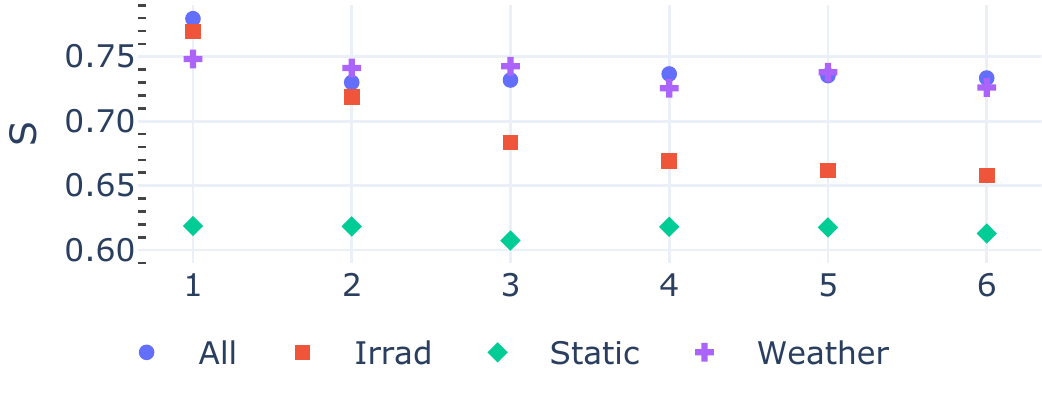}
    \caption{\ac{Serror} per step - higher is better}
    \label{fig:scatter_4mode_inputs_skill_per_step_global}
\end{subfigure}
\caption{Distribution of error per step for each of the inputs}
\label{fig:scatter_4mode_dnn_inputs_*_per_step}
\end{figure}

\begin{figure}[ht!]
    \begin{subfigure}[b]{0.5\textwidth}
    \centering
        \includegraphics[width=\textwidth]{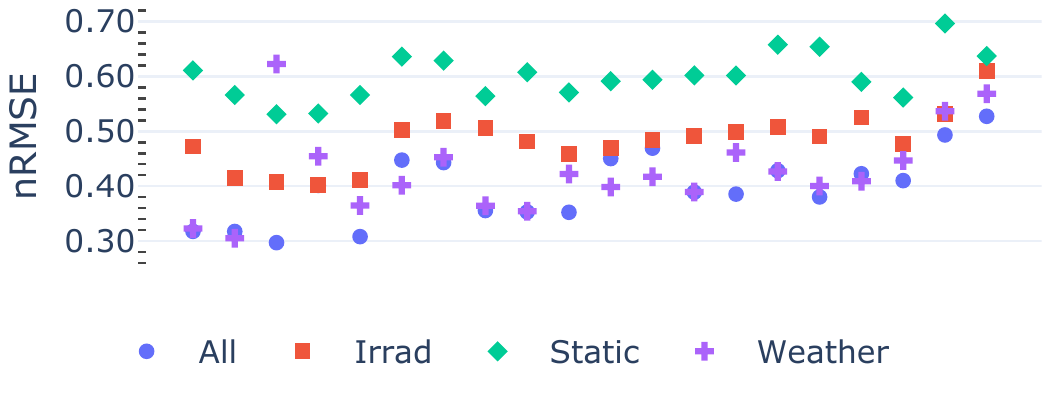}
        \caption{Local \ac{nRMSE} per \ac{AOI} - lower is better}
        \label{fig:scatter_4mode_inputs_nRMSE_per_plant_local}
    \end{subfigure}
    \begin{subfigure}[b]{0.5\textwidth}
    \centering
        \includegraphics[width=\textwidth]{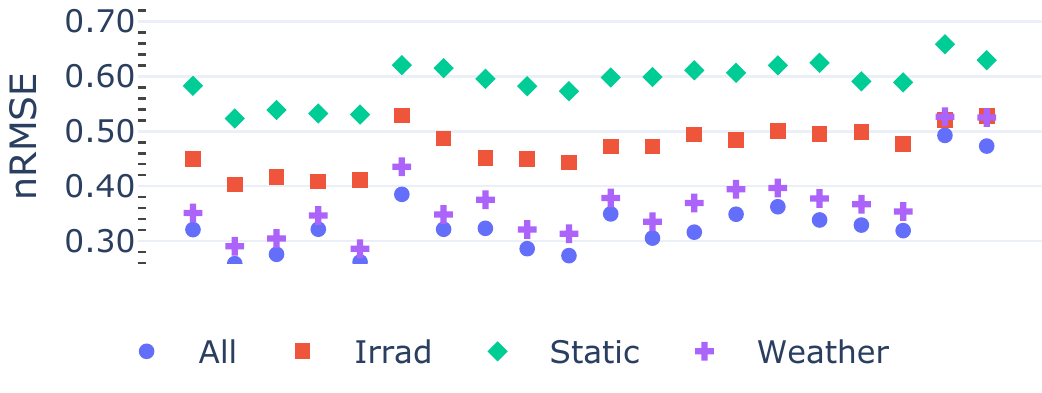}
        \caption{Global \ac{nRMSE} per \ac{AOI} - lower is better}
        \label{fig:scatter_4mode_inputs_nRMSE_per_plant_global}
    \end{subfigure}
    \begin{subfigure}[b]{0.5\textwidth}
    \centering
        \includegraphics[width=\textwidth]{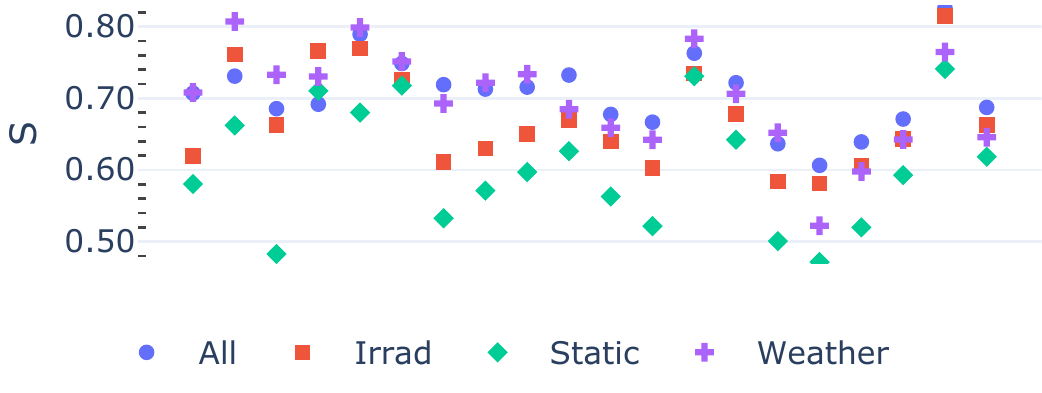}
        \caption{\ac{Serror} per \ac{AOI} - higher is better}
        \label{fig:scatter_4mode_inputs_skill_per_plant_local}
    \end{subfigure}
    \begin{subfigure}[b]{0.5\textwidth}
    \centering
        \includegraphics[width=\textwidth]{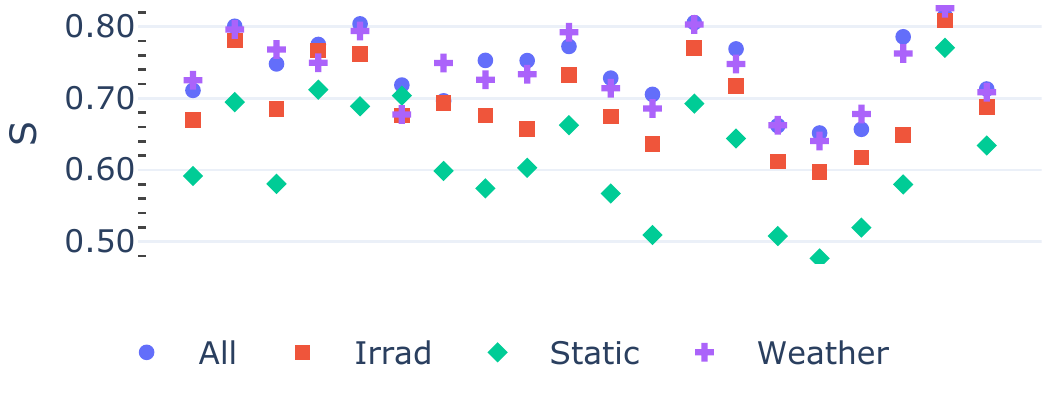}
        \caption{\ac{Serror} per \ac{AOI} - higher is better}
        \label{fig:scatter_4mode_inputs_skill_per_plant_global}
    \end{subfigure}
\caption{Distribution of error per \ac{AOI} for each of the inputs}
\label{fig:scatter_4mode_dnn_inputs_*_per_plant}
\end{figure}

In \autoref{tab:dnn_inputs_overview} we present the average test results across all \acp{AOI} and every forecast step (1 - 6).
We boldface the overall best Local and Global results for both the \ac{nRMSE} and \ac{Serror}.
We further break down the results by averaging the \ac{nRMSE} and \ac{Serror} per step in \autoref{fig:scatter_4mode_dnn_inputs_*_per_step} and per \ac{AOI} in \autoref{fig:scatter_4mode_dnn_inputs_*_per_plant}.
Additionally, \autoref{tab:friedman_inputs} shows the results of the Friedman test ranking the input combinations at forecast steps one and six.

\begin{table}[!ht]
    \centering
    \begin{tblr}{lcc |[dashed] cc |[2pt] cc |[dashed] cc}
    \toprule
    & \SetCell[c=4]{c} \textbf{\ac{nRMSE}}    &&&& \SetCell[c=4]{c} \textbf{\ac{Serror}} \\
    
    & \SetCell[c=2]{c} \textbf{Local} && \SetCell[c=2]{c} \textbf{Global} && \SetCell[c=2]{c} \textbf{Local} && \SetCell[c=2]{c} \textbf{Global}  \\
    & mean & std & mean & std & mean & std & mean & std \\ 
    \midrule
All     & \textbf{0.397} & 0.074 & \textbf{0.333} & 0.067 & \textbf{0.706} & 0.055 & \textbf{0.741} & 0.055 \\
Irradiance   & 0.483 & 0.079 & 0.470 & 0.086 & 0.671 & 0.077 & 0.693 & 0.073 \\
Static  & 0.600 & 0.043 & 0.591 & 0.036 & 0.603 & 0.083 & 0.616 & 0.079 \\
Weather & 0.426 & 0.088 & 0.370 & 0.066 & 0.699 & 0.072 & 0.737 & 0.053
    \end{tblr}
    \centering
    \caption{Average of \ac{nRMSE} (lower is better) and \ac{Serror} 
(higher is better) across all \acp{AOI} and forecast steps showing the effect of using various input features with a \ac{DNN}.}
    \label{tab:dnn_inputs_overview}
\end{table}

\begin{table}[!htp]
\centering
\resizebox{\textwidth}{!}{
 \begin{tblr}{ll cc|[dashed]cc |[2pt] cc|[dashed]cc}
 \toprule
      && \SetCell[c=4]{c} \textbf{{Local}} &&&& \SetCell[c=4]{c} \textbf{{Global}} \\
    && \SetCell[c=2]{c}\textbf{{STEP 1}}  && \SetCell[c=2]{c}\textbf{{STEP 6}} && \SetCell[c=2]{c}\textbf{{STEP 1}}    && \SetCell[c=2]{c}\textbf{{STEP 6}} \\
&& Rank & $p$ &  Rank & $p$  & Rank & $p$ & Rank & $p$ \\
    \midrule
\SetCell[r=4]{c}{\ac{nRMSE}}
 & All & \textbf{1.5} & - & \textbf{1.5} & - & \textbf{1.0} & - & \textbf{1.0} & - \\
 & Irradiance & 2.0 & 0.178 & 3.0 & 0.000 & 2.4 & 0.001 & 3.0 & 0.000 \\
 & Static & 3.9 & 0.000 & 3.9 & 0.000 & 4.0 & 0.000 & 4.0 & 0.000 \\
 & Weather & 2.6 & 0.020 & 1.7 & 0.713 & 2.6 & 0.000 & 2.0 & 0.010 \\
\hline
\SetCell[r=4]{c}{\ac{Serror}}
&All & 1.8 & 0.902 & \textbf{1.7}  & 0.903 & \textbf{1.4}  & -  &\textbf{1.4}  &  -    \\
&Irradiance & \textbf{1.7} &  - & 2.9  & 0.002 & 2.0 & 0.178 & 3.0    &  0.000     \\
&Static     & 4.0   & 0.000 & 3.8 &  0.000  & 4.0 &  0.000  & 3.9  & 0.000    \\
&Weather    & 2.5 & 0.002 & \textbf{1.7} &  -    & 2.7  & 0.001 & 1.7  & 0.462 \\
\end{tblr}
}
\caption{Average Rankings (Friedman) of the different inputs at steps one and six. Bold face is the best-ranked value per step} 
\label{tab:friedman_inputs}
\end{table}

From \autoref{tab:dnn_inputs_overview} we can see that use of just the static data produced the worse results by a significant margin. 
This is supported by the results in \autoref{tab:friedman_inputs} with static consistently ranking last.
Using all the inputs produced the best average results in all cases.
Furthermore, the use of all inputs either ranked best or had no significant difference to the best-ranked method in all cases.

In the case of using either just weather or just irradiance as additional data improves over the static only indicating that both features contain useful information the model can extract.
Looking at the per step errors in \autoref{fig:scatter_4mode_dnn_inputs_*_per_step} it is clear that when using just irradiance the largest effect on performance for the in the first few steps before dropping towards that of the static only.
Given a longer forecast horizon, we would expect this trend to continue until it eventually plateaus in line with the static as the importance irradiance decreases.
Conversely, the use of just weather data produces a consistent improvement at every step when compared to static only. 
This is unsurprising as, much like when using just static data, the information available for the model to produce the forecasts is the same for all steps.

While the models that use just irradiance outperform just weather at step one, they reach an inflexion point around step two or three.
This explains the switching of rankings at steps one and six in \autoref{tab:friedman_inputs}
When the model uses both irradiance and weather data, the behaviour is similar to that of just irradiance, starting at an improved state before dropping off, but only to be in line with that of the model that made use of weather data. 

Looking at \ac{nRMSE} errors the per \ac{AOI} in \autoref{fig:scatter_4mode_inputs_nRMSE_per_plant_global} there is a strata with relatively consistent performance increases for each of the input groups for all locations. 
However, this trend does not apply to the \ac{Serror} as can be seen in \autoref{fig:scatter_4mode_inputs_skill_per_plant_global}. 
This would indicate that improvement is not consistent relative to the forecast difficulty of each \ac{AOI}.
Looking at the \ac{Serror} per \ac{AOI} for the static input, given the fairly consistent \ac{nRMSE} values, the high variance would suggest that some locations are more challenging to predict for than others.
This gives more evidence that the inclusion of both weather and irradiance produces better models as the std of the \ac{Serror} drops from $0.08$ to $0.06$

Given these results, we can see that use of all features produces the best model.
It is worth noting that in both the Local and Global approaches use of just weather data appeared to be competitive only being clearly beaten in the first few time steps. This is especially evident when we look at the average forecast skill with both Local and Global having a difference of less than $0.01$.

\subsection{Local vs Global}\label{ch3:results:results_local_vs_global}

Here we compare the performance of the various learning methods trained using both the Local and Global data models.
We aim to understand and compare the Local and Global approaches.
We use all input features as, based on the results from the previous section, this produces the best models.

\begin{figure}[ht]
    \begin{subfigure}{0.5\textwidth}
        \includegraphics[width=\textwidth]{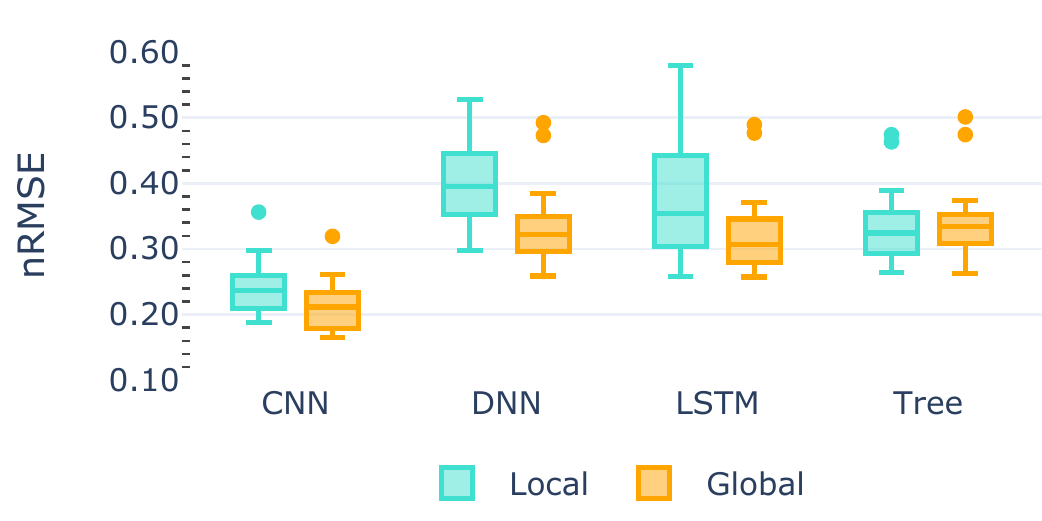}
        \caption{\ac{nRMSE} - lower is better}
    \end{subfigure}
    \begin{subfigure}{0.5\textwidth}
        \includegraphics[width=\textwidth]{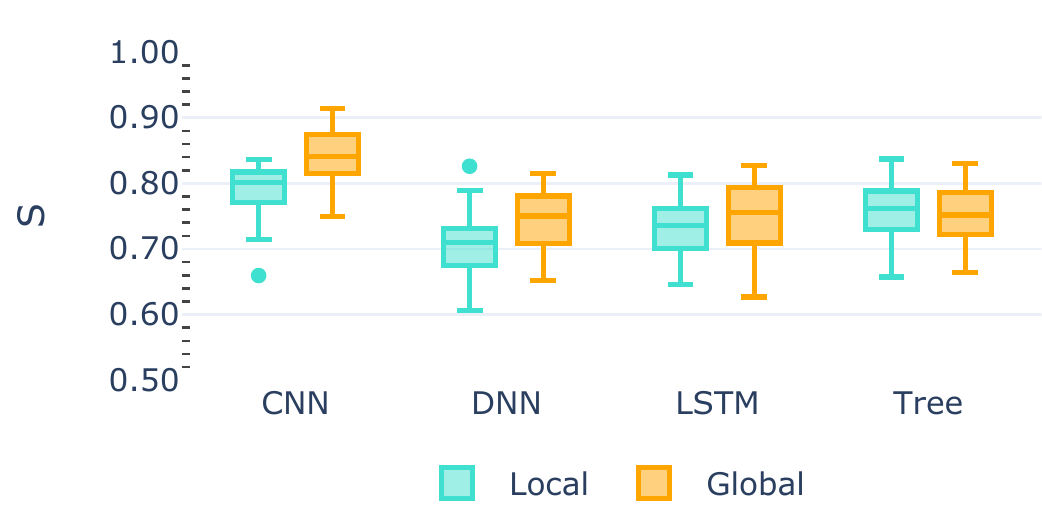}
        \caption{\ac{Serror} - higher is better}
    \end{subfigure}
    
    \caption{Distribution of Local and Global errors}
    \label{fig:boxplot_grouped_lg_rmse_skill}
\end{figure}

\begin{figure}[ht]

\begin{subfigure}[]{0.5\textwidth}
    \includegraphics[width=\textwidth]{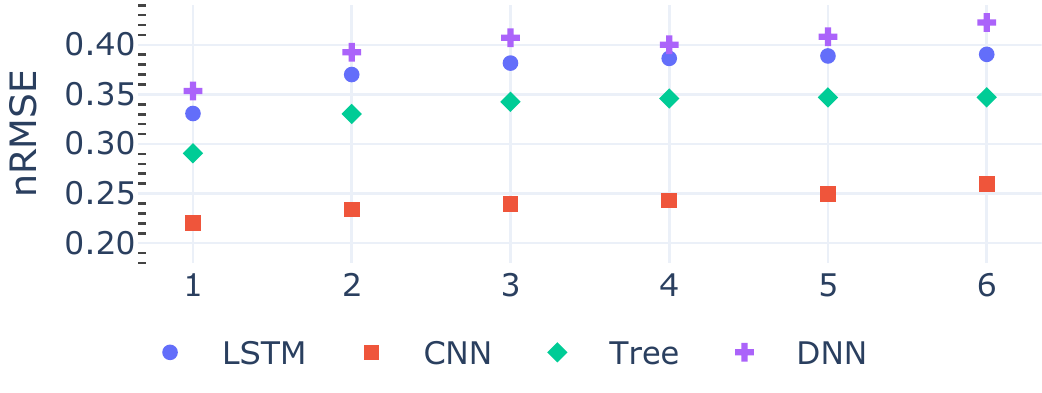}
    \caption{\ac{nRMSE} per step for Local - lower is better}
    \label{fig:scatter_4mode_nRMSE_per_step}
\end{subfigure}
\begin{subfigure}[]{0.5\textwidth}
    \includegraphics[width=\textwidth]{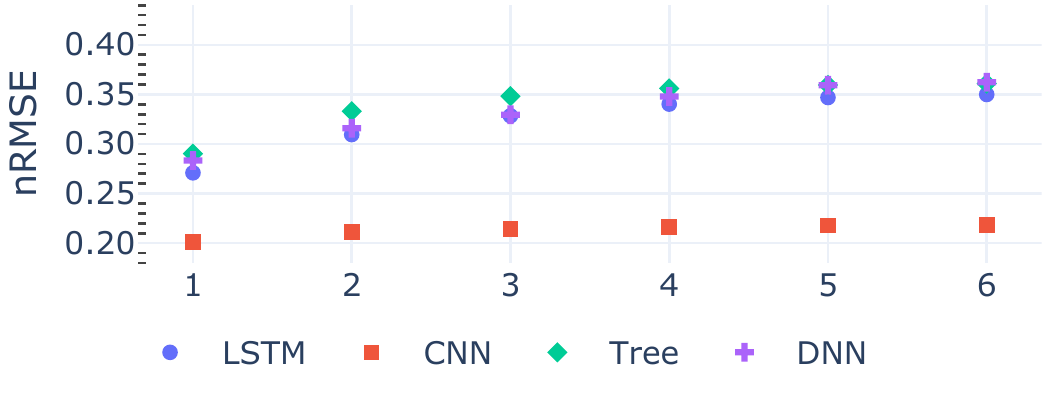}
    \caption{\ac{nRMSE} per step for Global - lower is better}
\end{subfigure}
\begin{subfigure}[]{0.5\textwidth}
    \includegraphics[width=\textwidth]{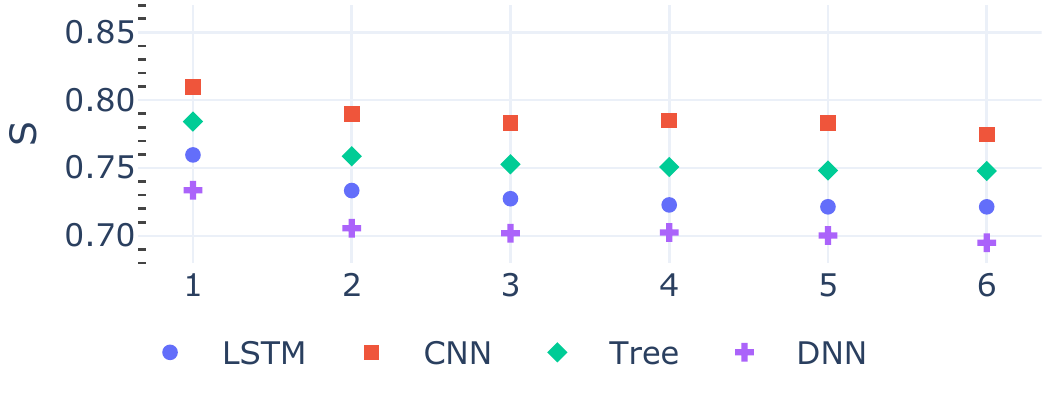}
    \caption{\ac{Serror} per step for Local - higher is better}
    \label{fig:scatter_4mode_skill_per_step}
\end{subfigure}
\begin{subfigure}[]{0.5\textwidth}
    \includegraphics[width=\textwidth]{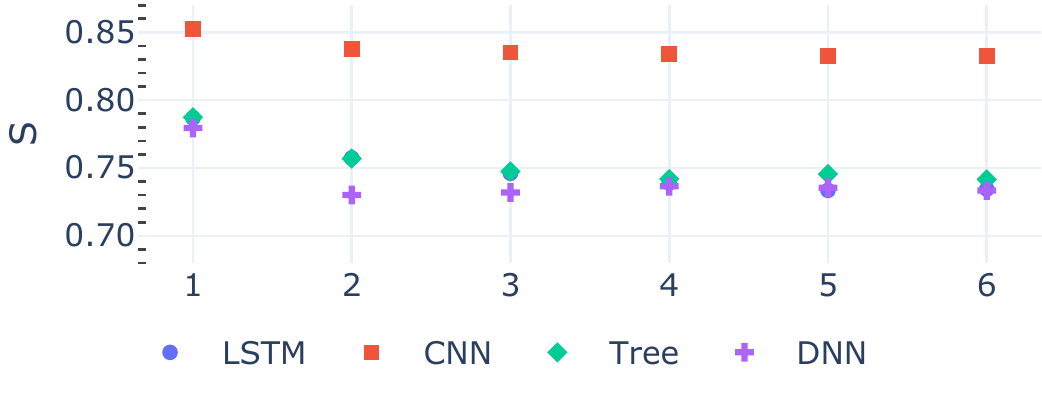}
    \caption{\ac{Serror} per step for Global - higher is better}
\end{subfigure}
\caption{Distribution of error per step for each of the models comparing local and global}
\label{fig:scatter_4mode_all}
\end{figure}

\autoref{table:ch3:overview_local_global} gives an overview of the \ac{nRMSE} and \ac{Serror} results for each model. 
We present both the overall average for all \acp{AOI} at every forecast step, 1-6, as well as the standard deviation. 
In italics, we emphasise if the Local or Global approach produced the best result for each model while in boldface is the best overall model.
\autoref{fig:boxplot_grouped_lg_rmse_skill} and \autoref{fig:scatter_4mode_all} show the spread of the \ac{nRMSE} and \ac{Serror} for each \ac{AOI} and forecast step per model. 

Looking at the average results in \autoref{table:ch3:overview_local_global} it is clear that the Global approach outperforms the Local in all the \ac{DL} methods (\ac{CNN}, \ac{DNN}, \ac{LSTM}). However, for the Trees, performance is almost identical between the two approaches with the Local flavour presenting a slightly better average. 
At steps one and six this difference is insignificant with a $p$ value $> 0.2$.

Looking at the per step errors in \autoref{fig:scatter_4mode_all} we see that the Global approach improves results at every step for the \ac{DL} methods.
We can see from \autoref{fig:boxplot_grouped_lg_rmse_skill} that for the CNN and DNN in addition to improving mean error, there is a lower variance per \ac{AOI} when using the Global method.
We suspect this is because the Global models are able to extract information from one \ac{AOI} and apply it to another.


\begin{table}[ht!]
    \centering
    \begin{tblr}{lcc |[dashed] cc |[2pt] cc |[dashed] cc}
    \toprule
    & \SetCell[c=4]{c} \textbf{\ac{nRMSE}}    &&&& \SetCell[c=4]{c} \textbf{\ac{Serror}} \\
    
    & \SetCell[c=2]{c} \textbf{Local} && \SetCell[c=2]{c} \textbf{Global} && \SetCell[c=2]{c} \textbf{Local} && \SetCell[c=2]{c} \textbf{Global}  \\
    & mean & std & mean & std & mean & std & mean & std \\ 
    \midrule
  CNN & 0.241 & 0.042 &  \textbf{\textit{0.213} }& 0.038 & 0.788 & 0.048 &  \textbf{\textit{0.837}} & 0.043 \\
  DNN & 0.397 & 0.074 & \textit{ 0.333 }& 0.067 & 0.706 & 0.055 &  \textit{0.741} & 0.055 \\
 LSTM & 0.375 & 0.090 &  \textit{0.324} & 0.069 & 0.731 & 0.047 &  \textit{0.750} & 0.059 \\
 Tree & \textit{0.334} & 0.060 &  0.341 & 0.063 & \textit{0.757} & 0.056 &  0.753 & 0.053 \\

\end{tblr}

\caption{Average of Global and Local results for steps 1-6 for each model. In italic is the best of either Local or Global for each method and metric. Bold face is the best overall method for the metric.}
\label{table:ch3:overview_local_global}

\end{table}

\begin{table}[!htp]
    \centering
    \begin{tblr}{c c c c c c c}
    \toprule
    Model&Step 1 & Step 2 & Step 3 & Step 4 & Step 5 & Step 6\\
    \midrule
    CNN& 1.0& 1.0& 1.0& 1.0& 1.0& 1.0\\
    DNN&3.5 &3.8 &3.5 &3.2 & 3.2 & 3.2\\
    LSTM&2.7 &2.5 &2.8 & 3 & 3.3 & 3.1\\
    Tree&2.9 &2.8 &2.8 & 2.9 & 2.6 & 2.7\\
    
    \end{tblr}
    \caption{Friedman Rankings of the S error for each Global model per step}
    \label{table:model_rankings}
\end{table}

\subsubsection{Does location affect performance?}

\begin{figure}
\begin{subfigure}[]{0.5\textwidth}
    \includegraphics[width=\textwidth]{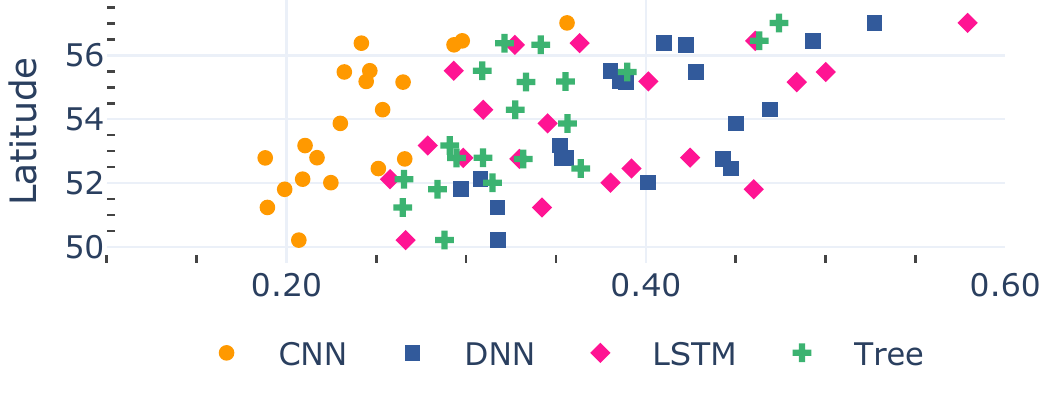}
    \caption{\ac{nRMSE} Local - lower is better}
\end{subfigure}
\begin{subfigure}[]{0.5\textwidth}
    \includegraphics[width=\textwidth]{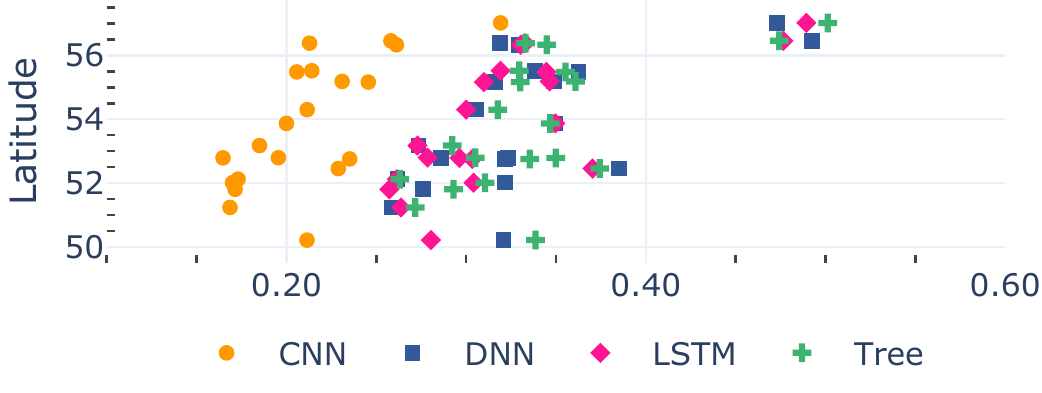}
    \caption{\ac{nRMSE} Global - lower is better}
\end{subfigure}

\begin{subfigure}[]{0.5\textwidth}
    \includegraphics[width=\textwidth]{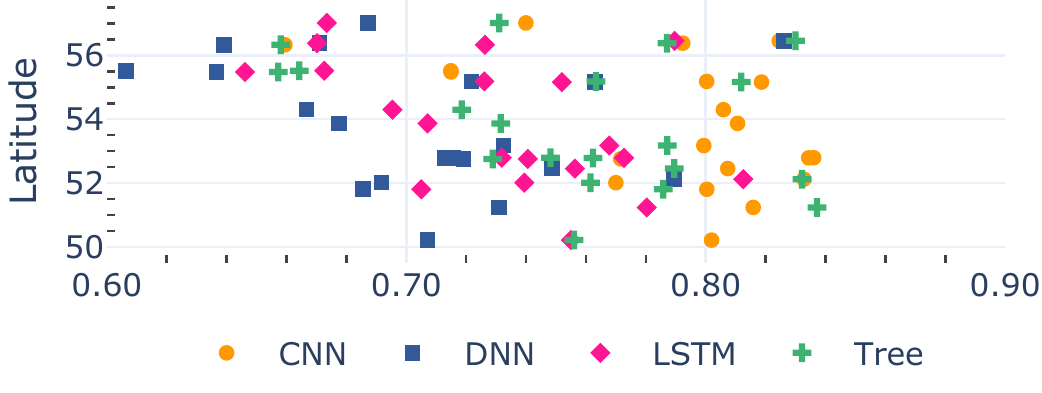}
    \caption{\ac{Serror} Local - higher is better}
\end{subfigure}
\begin{subfigure}[]{0.5\textwidth}
    \includegraphics[width=\textwidth]{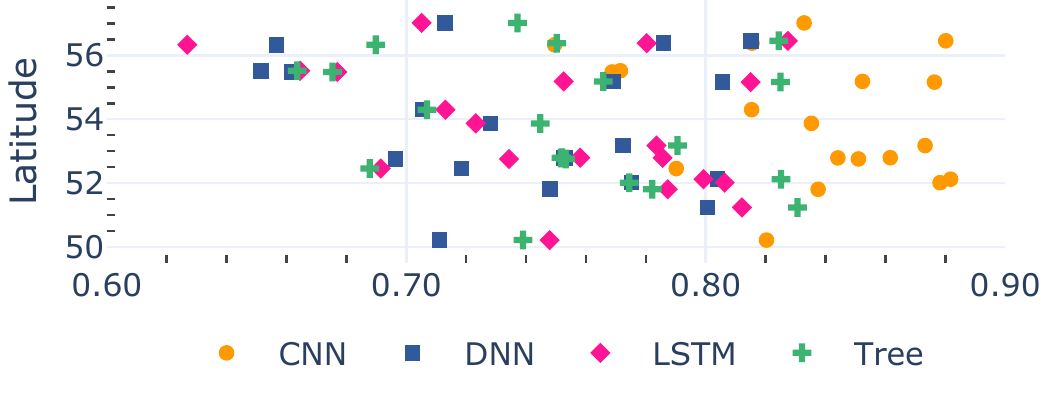}
    \caption{\ac{Serror} Global - higher is better}
\end{subfigure}
    \caption{Error for all time steps 1-6 compared to \ac{AOI} latitude (N/S)}
    \label{fig:scatter_lg_rmseS_location}
\end{figure}

In \autoref{fig:scatter_lg_rmseS_location} we show the error for every time step per \ac{AOI} plotted against the \acp{AOI} latitude. This gives us an indication if there is any correlation between how far North / South an \ac{AOI} is and its performance relative to the other \acp{AOI}. 

When looking at the \ac{nRMSE} there is a correlation between the \acp{AOI} error and its latitude with \acp{AOI} further north appearing to perform worse.
$r = 0.72$ and $r = 0.68$ for Local and Global respectively.
However, when comparing the \ac{Serror} error, the correlation is not as strong, 
with Local $r = -0.43$ and Global $r = -0.30$.
We believe this is because the \ac{nRMSE} values are normalised by the mean irradiance of the \ac{AOI} and locations further north receive less irradiance throughout the year and as such have a lower average exacerbating any forecast error relative to \ac{AOI}s further south.

Overall, while location likely does play some part in the performance of any given 
\ac{AOI}, we feel it is not as significant a factor relative to other factors that may affect model performance at any \ac{AOI}.

\subsubsection{Data Used / Learning methods}
Of the methods that make use of ground-based point weather; DNN, LSTM and Trees, their performance is extremely consistent. 
This is especially true for the Global LSTMs and DNNs where their performance is almost identical as can be seen in \autoref{fig:scatter_4mode_all}. 
It is also emphasised in \autoref{table:model_rankings} showing the Global model rankings at every step.
The fact that all three methods appear to perform comparably while the CNNs show a significant improvement, could suggest that there is a limit on the amount of information that can be extracted using ground-based weather for the forecasts. 
The use of the image data seems to break through the information floor, as is supported by the fact that even the local CNN outperforms all other methods' Global approach. 

We suspect this is due to the fact it receives a richer view of the weather as there is the distance between the observation station and the \ac{AOI}.
However, this improved performance comes with a much larger computational cost to train and run the models.
\autoref{tab:approx_training_times} gives an overview of the approximate training times for each method.
The \acp{CNN} takes significantly longer to train than the other methods.

As the time complexity for all the methods used is $O(n)$, sequentially training a Local model for all \ac{AOI}s or a single Global model will take approximately the same amount of time.
Of course, in practice, it is possible to easily parallelise training the local approach\footnote{All \ac{DL} Local models were trained using vector parallelism saving significant time}.
This is not possible for the Global approach as the model needs to be trained on all the data.
This is also one of the advantages of the Global approach, and we suspect a reason it outperforms the Local, it sees more data.

The same training time constraints are true for the Trees as their training time grows in line with the amount of data you need to learn from. However, the Global approach does not seem to provide a performance gain.


\begin{table}[h]
    \centering
    \begin{tblr}{c|c|c|c}
    \toprule
        Model & Local & Global  & Compute \\ 
    \midrule
        CNN   & 1 hour & 20 hours & 1x A6000 GPUs \\
        DNN  &  2 mins & 20 mins & 2x 2080 GPUs \\
LSTM  & 3 mins & 30 mins & 2x 2080 GPUs \\
Trees &  24 CPU mins & 500 cup mins & local 2 cores, global 10 cores 
    \end{tblr}
    \caption{Approximate run times to train the models. 
    For the trees, this is the time to train 6 trees, one for each forecast step.
    A large part of the Local training time (~30\%) is spent on overheads such as filling caches etc. A single Local training epoch for the DNN took approximately 5 seconds.}
    \label{tab:approx_training_times}
\end{table}

\subsubsection{Conclusion}
When working with multiple \acp{AOI}, the Global approach is better for \ac{DL} based methods.
While they take longer to train, they produce better and more consistent results.
Additionally, in the case of the \ac{DL} approaches if new data become available they could potentially be further refined by training the existing model on the new data transfer learning.
When there are only a few \ac{AOI} Trees, there is a minimal performance gain compared to using the local version.

\subsection{\Ac{cv-mode} and \Ac{kn-mode}}\label{ch3:results:cv_kn}
We have shown in \autoref{ch3:subsection:effect_of_input}, the use of real-time irradiance can improve accuracy for the first few steps of the predictions.
However, until now we have presented results for the perfect case where data has been available for all \acp{AOI}.
One of the main aims of our paper is to present an understanding of potential solutions for when there is no or limited access to data at the \ac{AOI}.
In \autoref{ch3:subsection:data_models} we presented two data models able to produce forecasts for \acp{AOI} with limited data, \ac{cv-mode} for a lack of historic data and \ac{kn-mode} for a lack of real-time data at the \ac{AOI}.
In this Section, we analyse the performance of models trained using these alternate data models compared to the Local and Global approaches.

\begin{figure}[t]
\begin{subfigure}{0.5\textwidth}
    \centering
    \includegraphics[width=7cm]{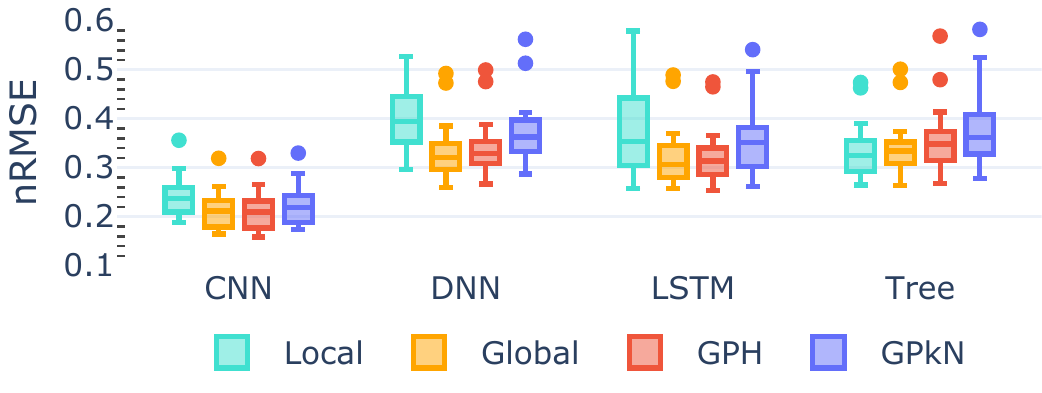}
    \caption{\ac{nRMSE} - lower is better}
\end{subfigure}
\begin{subfigure}{0.5\textwidth}
        \centering
        \includegraphics[width=7cm]{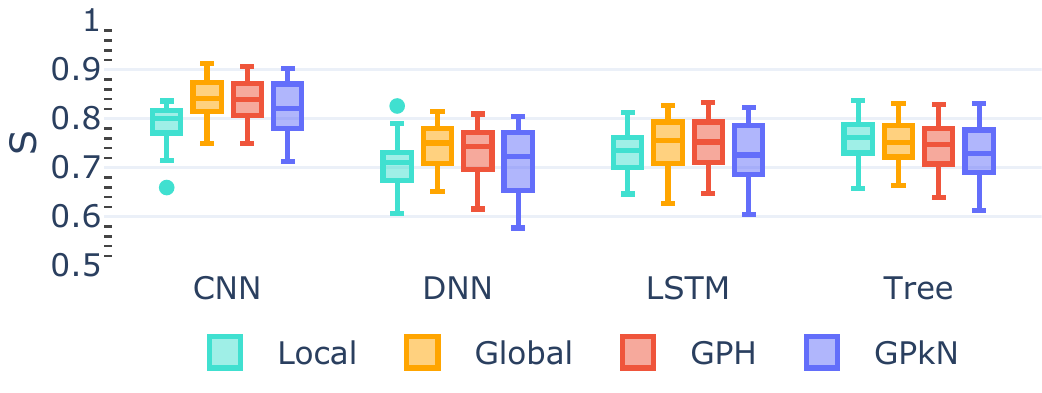}
        \caption{\ac{Serror} - higher is better}
\end{subfigure}
    
    \caption{Overview of Local, Global, \ac{cv-mode} and \ac{kn-mode} per \ac{AOI}. The Box plot shows the average error for each \ac{AOI}}
    \label{fig:boxplot_grouped_lgck_rmse_skill}
\end{figure}

\begin{figure}
    \begin{subfigure}{0.5\textwidth}
        \centering
        \includegraphics[width=7cm]{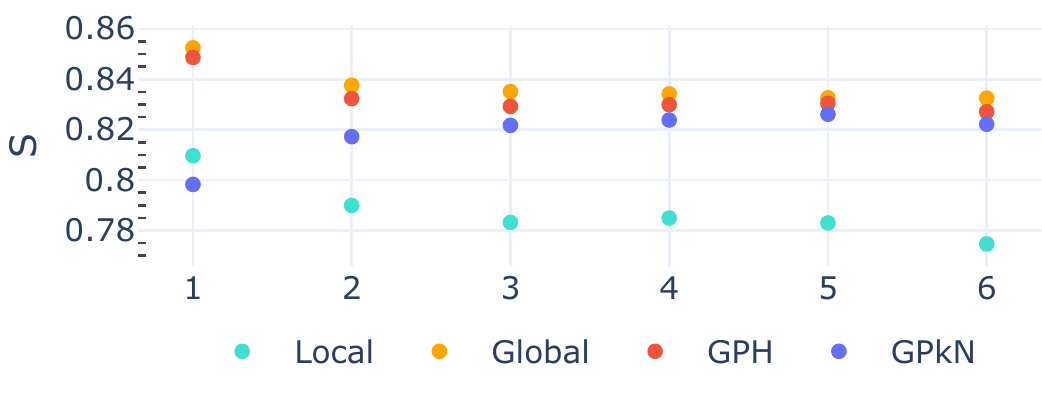}
        \caption{\ac{Serror} per step for CNN}
        \label{fig:scatter_CNN_lgck_skill}
    \end{subfigure}
    \begin{subfigure}{0.5\textwidth}
        \centering
        \includegraphics[width=7cm]{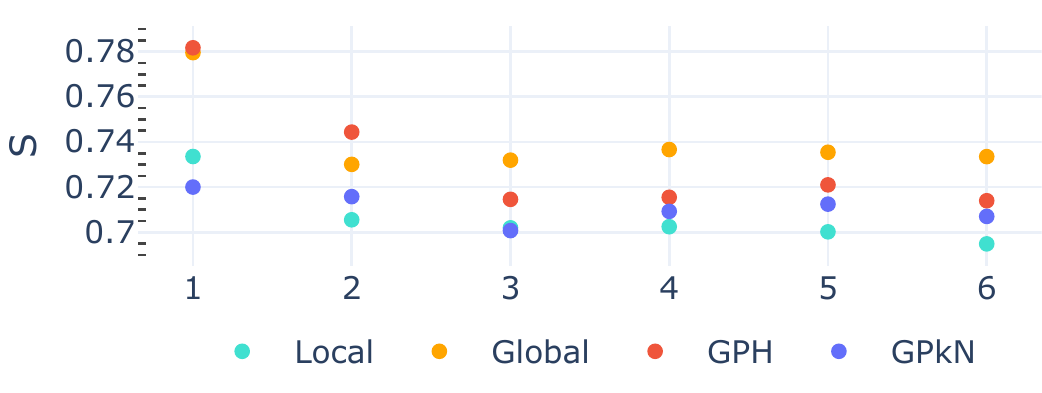}
        \caption{\ac{Serror} per step for DNN}
        \label{fig:scatter_DNN_lgck_skill}
    \end{subfigure}
    \begin{subfigure}{0.5\textwidth}
        \centering
        \includegraphics[width=7cm]{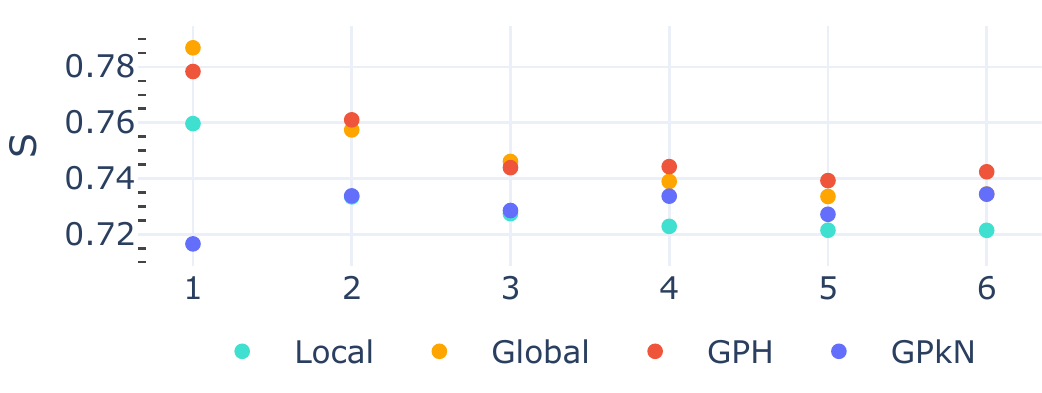}
        \caption{\ac{Serror} per step for LSTM}
        \label{fig:scatter_LSTM_lgck_skill}
    \end{subfigure}
    \begin{subfigure}{0.5\textwidth}
        \centering
        \includegraphics[width=7cm]{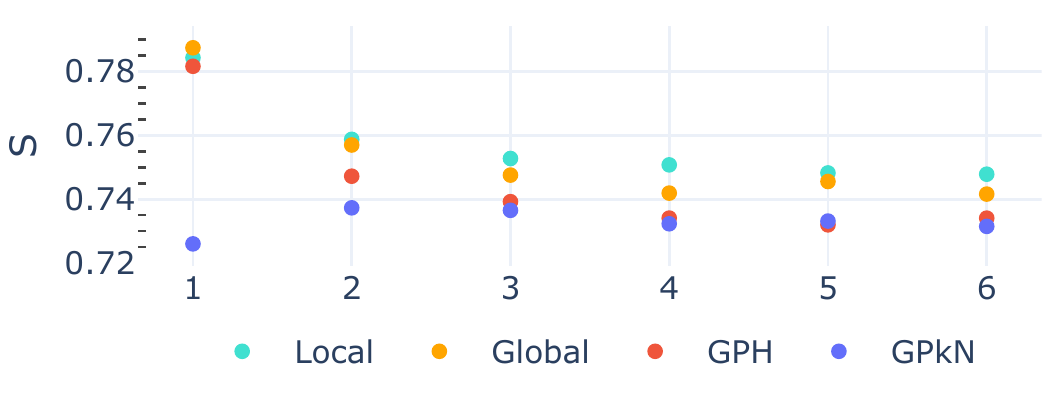}
        \caption{\ac{Serror} per step for Tree}
        \label{fig:scatter_Tree_lgck_skill}
    \end{subfigure}
    
    \caption{\ac{Serror} error per forecast step for the Local, Global, \ac{cv-mode} and \ac{kn-mode} approach for at steps 1-6 for each of the 4 models}
    \label{fig:scatter_per_step_*_lgck_skill}
\end{figure}

In \autoref{fig:boxplot_grouped_lgck_rmse_skill} we show the distribution of error per \ac{AOI} for each of the models trained using all four data models, Local, Global, \ac{cv-mode} and \ac{kn-mode}. \autoref{fig:scatter_per_step_*_lgck_skill} shows a breakdown of the \ac{Serror} for each of the four models at every forecast step.
While \autoref{tab:lgck_ranks} shows the Friedman ranking for each method at forecast steps one and six.

\begin{table}[t]
\centering
   \caption{Friedman rankings of the different data models at steps one and six for both error metrics. The best-ranked method is in bold}
   \label{tab:lgck_ranks}
    
    \resizebox{\textwidth}{!}{
    \begin{tblr}{ll cc |[dashed] cc |[2pt] cc |[dashed] cc}
    \toprule
 & & \SetCell[c=4]{c}Step 1 &&&& \SetCell[c=4]{c}Step 6 &&&  \\
&  & \SetCell[c=2]{c}RMSE && \SetCell[c=2]{c}S && \SetCell[c=2]{c}RMSE && \SetCell[c=2]{c} S &  \\
 &  & Ranking & $p$ & Ranking & $p$ & Ranking & $p$ & Ranking & $p$ \\
 \midrule
\SetCell[r=4]{c}{CNN} 
 & CV     & 1.7 & 0.462 & \textbf{1.7} & - & \textbf{1.7} & - & \textbf{2.0} & - \\
 & Global & \textbf{1.4} & - & \textbf{1.7} & 1.000 & 2.2 & 0.178 & 2.1 & 0.903 \\
 & KN     & 3.8 & 0.000 & 3.4 & 0.000       & 2.2 & 0.221 & 2.2 & 0.624 \\
 & Local  & 3.2 & 0.000 & 3.3 & 0.000       & 4.0 & 0.000 & 3.8 & 0.000 \\
 \hline
\SetCell[r=4]{c}{DNN}
 & CV     & 1.8          & 0.142 & \textbf{1.6} & -     & 2.3          & 0.002 & 2.4          & 0.010 \\
 & Global & \textbf{1.2} & -     & 1.8          & 0.713 & \textbf{1.1} & -     & \textbf{1.3} & - \\
 & KN     & 3.7          & 0.000 & 3.4          & 0.000 & 3.1          & 0.000 & 3.0          & 0.000 \\
 & Local  & 3.3          & 0.000 & 3.3          & 0.000 & 3.6          & 0.000 & 3.4          & 0.000 \\
 \hline
\SetCell[r=4]{c}{LSTM} 
 & CV       & 2.0           & 0.037 & 2.0           & 0.327 & \textbf{1.8}  & -     & \textbf{2.0} & - \\
 & Global   & \textbf{1.2}  & -     & \textbf{1.6}  & -     & 2.7           & 0.043 & 2.7 & 0.173 \\
 & KN       & 3.7           & 0.000 & 3.6           & 0.000 & 2.8           & 0.043 & 2.6 & 0.142 \\
 & Local    & 3.2           & 0.000 & 2.8           & 0.007 & 2.8           & 0.043 & 2.9 & 0.082 \\
\hline
\SetCell[r=4]{c}{Tree} 
& CV     & 2.5          & 0.014 & 2.5          & 0.132  & 3.3           & 0.000 & 2.9          & 0.259 \\
& Global & \textbf{1.5} & -     & \textbf{1.7} & -      & 1.9           & 0.391 & \textbf{2.2} & - \\
& KN     & 4.0          & 0.000 & 3.7          & 0.000  & 3.3           & 0.000 & 2.8          & 0.259\\
& Local  & 2.0          & 0.270 & 2.2          & 0.270  & \textbf{1.6}  & -     & 2.2          & 0.903
    \end{tblr}
    }
\end{table}

From \autoref{fig:boxplot_grouped_lgck_rmse_skill} it is clear that for the \ac{DL} methods both \ac{cv-mode} and \ac{kn-mode} appear to be viable approaches outperforming their Local counterparts.
The \ac{cv-mode} approach yields results in line with the Global approach.
Looking at the per-step errors in \autoref{fig:scatter_per_step_*_lgck_skill} it is clear they follow the same trend with a number of steps being indistinguishable.
This is supported by rankings in \autoref{tab:lgck_ranks} with Global and \ac{cv-mode} consistently ranking in line with one another.

In the case of \ac{kn-mode} applied to the \ac{DL} models from \autoref{fig:boxplot_grouped_lgck_rmse_skill} they appear to fit between the Local and Global methods.
Looking at the per-step error in \autoref{fig:scatter_per_step_*_lgck_skill} it is clear that this is the result of \ac{kn-mode} under-performing relative to even the Local approach for the first few forecast steps. 
The \ac{kn-mode} eventually rises to be only marginally worse than the \ac{cv-mode}.
We can see this effect in the rankings, \ac{kn-mode} ranks last at step one but by step six it usually ranks closer to the \ac{cv-mode} and always outranks Local.

The Trees, however, seem to buck this trend.
As already noted in \autoref{ch3:results:results_local_vs_global}, their performance for Local and Global data models has very little difference. 
Interestingly both the \ac{cv-mode} and \ac{kn-mode} versions seem to perform comparably and worse than both the Local and Global approaches. 
This indicates that the Trees fail to generalise to unseen locations.
This could suggest that they overly rely on correlations learned from the real-time irradiance of the \ac{AOI} seen to make their predictions.

Overall, both \ac{cv-mode} and \ac{kn-mode} appeared to be viable methods for generating forecasts for \ac{AOI}s where there is limited data available.

\subsubsection{Another look at input features}\label{ch3:results:cv_kn:inputs}
Both the \ac{cv-mode} and \ac{kn-mode} data models enable forecasts for \ac{AOI} without historic or direct access to data. 
However, since all the models use some value of observed irradiance, there is still a dependency on real-time data.
This is not always possible to access. 
In this section, we further analyse the effects the use of real-time irradiance has on performance compared to just using weather data.
We compare the results from the various models trained both with and without real-time irradiance as an input feature. We specifically focus on the results from the Local, Global and \ac{kn-mode} modes. \ac{cv-mode} was omitted as without irradiance as an input feature the results are the same as \ac{kn-mode}.

\begin{figure}[ht]
\begin{subfigure}{\textwidth}
\centering
        \includegraphics[width=\textwidth]{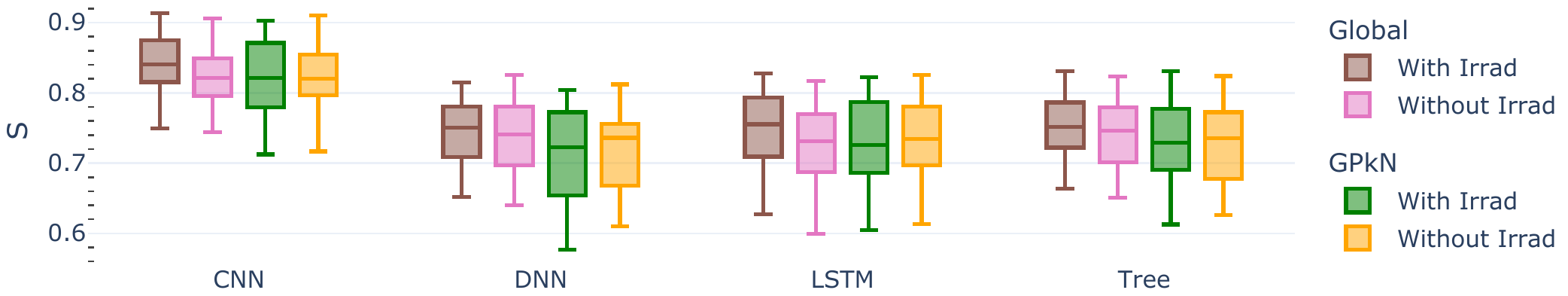}
        \caption{Distribution of the average \ac{Serror} for every AOI}
\end{subfigure}
\begin{subfigure}{\textwidth}
    \centering
    \includegraphics[width=\textwidth]{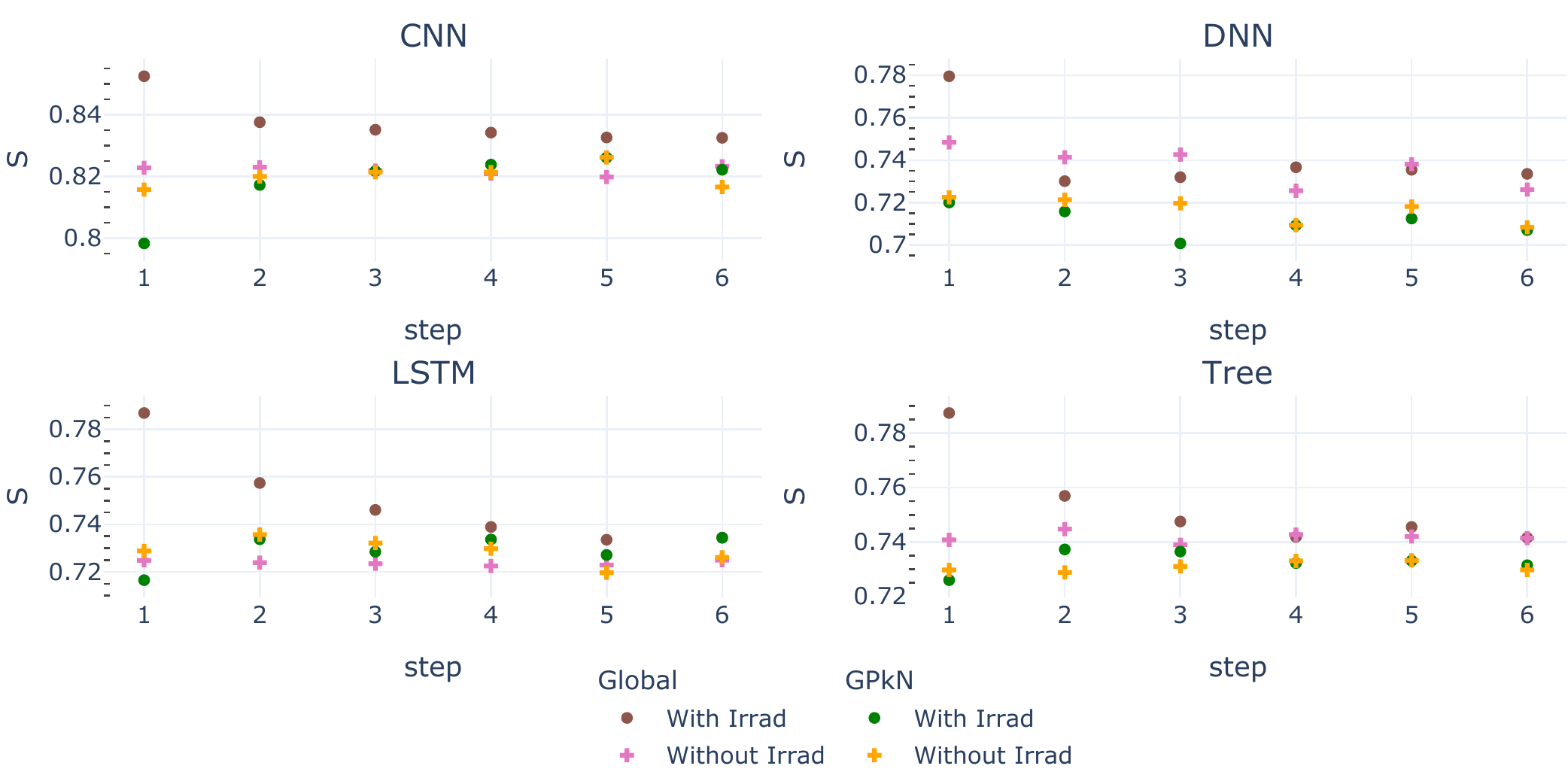}
    \caption{Average \ac{Serror} per step}
\end{subfigure}

\caption{A comparison of Global and \ac{kn-mode} for all the \ac{ML} methods with and without irradiance as in input used as an input feature}
\label{fig:gk_irrad_vs_noirrad_skill} 
\end{figure}

\autoref{fig:gk_irrad_vs_noirrad_skill} shows the average \ac{Serror} per \ac{AOI} and per step of the Global and \ac{kn-mode} models trained both with and without real-time irradiance as an input feature. 
In the case of Global models, we clearly see the inclusion of real-time irradiance improves performance for all the models.
This is unsurprising as we have already shown in \autoref{ch3:subsection:effect_of_input} that the use of irradiance helps Global models.
However, in the case of \ac{kn-mode} where each \ac{AOI} does not have access to real-time and instead uses a substitute value, there is not a clear performance benefit.

In fact, in the case \ac{CNN} performance appears to drop for the first time step when using the substituted irradiance. 
However, when performing the Wilcoxon test this difference was not significant with $p=0.13$ this would suggest a few \acp{AOI} may be skewing the results.
This trend held for all the \ac{kn-mode} models, unlike the Global models the difference when using irradiance is not statistically significant. In other words when operating in the worst-case scenario of \ac{kn-mode} the inclusion of irradiance has no real benefit.

We suspect this is due to a poor correlation between the true real-time irradiance used by the Global and the substituted value used in \ac{kn-mode}.
With the average distance to the substituted \ac{AOI} being 68km this is unsurprising.
If the substituted values were closer we would expect to see performance improve.

As we observed in \autoref{ch3:subsection:effect_of_input} for Global models, where the real-time irradiance is sauced from the \ac{AOI}, its inclusion improves model performance for the first few forecast step.
However, for \ac{kn-mode}, when irradiance is not sourced from the \ac{AOI}, the use of irradiance is of limited value and in some cases may even hinder model performance.
As such, when using \ac{kn-mode} models care must be taken to ensure a correlation of irradiance between the target \ac{AOI} and the substituted \ac{AOI}.

\subsection{Results Summary}
Overall Global models with the use of real-time irradiance perform the best. 
However, their need for a complete dataset can limit their usefulness in the real world.
We have shown through the use of \ac{cv-mode} that the Global models generalise well across locations working well for unseen locations.

More generally it would seem that the inclusion of real-time irradiance can improve overall performance for an \ac{AOI} where it is available, regardless of the data and training model used. 
However, its advantage over just using weather data seems to be limited to the first few forecast steps, and for forecasts with horizons longer than a few hours, its use is not as necessary.
We have shown it is possible to uncouple irradiance observations from forecasting models, although with marginally reduced performance.


\section{Conclusion}\label{section:conclusion}

In this work, we have explored various techniques for building irradiance forecasting models.
We used a number of standard \ac{ML} methods, RFs, \acp{DNN}, \acp{LSTM} and \acp{CNN}.
Each trained using four approaches: Local, Global, \acf{cv-mode} and \acf{kn-mode}.
The Local approach trains a model per location while the Global approach combines data from all locations and trains a single model.
\Ac{cv-mode} and \ac{kn-mode} are extensions to the Global approach used to circumvent data dependency issues that may occur at training time and when the model is in production. 
\ac{cv-mode} model generates forecasts for locations without historic data while \ac{kn-mode} circumvents any real-time data dependency by substituting values from nearby locations. 
Furthermore, we analysed the effects the use of diffident input features can have, specifically; real-time irradiance and weather data. 
We also explored different weather formats, point-based and satellite data.

Experimentally, we have shown that the Global approach and its extensions are superior to the Local.
While computationally more expensive to train, utilising all sequences to learn, the single Global model consistently outperformed its local counterpart.
Furthermore, the Global approach can be utilised to generate forecasts for locations with limited historical data.


Our experiments have shown that the use of real-time irradiance can improve forecasts for the first few steps, however, after 2-3 hours its importance diminishes, and weather data is key.
When attempting to forecast for locations without direct access to real-time data, while it is possible to substitute irradiance values from other locations care must be taken. 
The greater the distance between the two locations, the weaker their irradiance will correlate, and performance will be negatively impacted.

Additionally, our experimental results have shown that the use of satellite images works very well.
While the RF, \ac{DNN}, and \ac{LSTM} perform in line with each other, the \ac{CNN} using satellite imagery consistently outperforms all of them.
In fact, the \ac{CNN} operating in its worst case, \ac{kn-mode}, presents a significant improvement over its Global ground-based weather counterparts.
While in practice, access to these kinds of forecasts may be challenging; this result would strongly suggest that the use of richer weather data, over single-point data, can significantly improve forecast accuracy. 

We feel that these models can be integrated into a planning and optimisation system for use in the energy market.
However, further exploration of the effect the use of richer weather data has on performance is needed. We plan work to combine satellite and ground-based weather data and increase the temporal resolution to better capture intra-hour fluctuations.
We also plan to extend the substitution approach used in \ac{kn-mode} by combining data from multiple locations rather than just using the closest location. 

\section{Acknowledgments}
The authors would like to thank Elastacloud for providing access to data and computing resources.
We gratefully acknowledge the support of NVIDIA Corporation with the donation of the Titan Xp GPU used for this research.
T. Cargan holds a studentship funded by EPSRC and The University of Nottingham.
I. Triguero is funded by a Maria Zambrano Senior Fellowship at the University of Granada. This work is also supported by the Spanish projects
A-TIC-434-UGR20 and PID2020-119478GB-I00.

\bibliographystyle{elsarticle-num} 
\bibliography{references}

\end{document}